\PassOptionsToPackage{table,dvipsnames}{xcolor}
 \documentclass[final,3p,times,twocolumn]{elsarticle}
\usepackage{array}
\usepackage{colortbl}
\usepackage{color}
\usepackage[table]{xcolor}
\usepackage{subfig}
\usepackage{caption}
\usepackage{lineno,hyperref}
\usepackage{enumitem}
\usepackage{graphicx}
\usepackage{algorithm}
\usepackage{algorithmic}
\usepackage{multirow}
\usepackage{extarrows}
\usepackage{epstopdf}
\usepackage{url}
\usepackage{multicol}
\usepackage{amssymb}
\usepackage{booktabs}
\usepackage{calc}
\usepackage{amsmath,mathtools}
\usepackage{braket}
\usepackage{arydshln}
\usepackage{utfsym}
\usepackage{bbding}
\usepackage{float}
\usepackage{graphicx}
\usepackage{fontawesome}
\usepackage{geometry}
\usepackage{longtable}
\usepackage{ltablex}
\usepackage{CJKutf8}
\modulolinenumbers[5]

\allowdisplaybreaks[4]
\journal{INFORMATION PROCESSING \& MANAGEMENT}

\begin{document}
\begin{CJK}{UTF8}{gkai}
\begin{frontmatter}

\title{SarcasmBench: Towards Evaluating Large Language Models on Sarcasm Understanding}


\author[tiandaaddress,polyuaddress]{Yazhou Zhang}
\author[zzuliaddress]{Chunwang Zou}
\author[zkyaddress]{Zheng Lian\corref{mycorrespondingauthor}}
\author[ruidianaddress3]{Prayag Tiwari\corref{mycorrespondingauthor}}
\author[polyuaddress]{Jing Qin}

\cortext[mycorrespondingauthor]{Corresponding authors.}

\address[tiandaaddress]{College of Intelligence and Computing, Tianjin University, Tianjin, China}
\address[zzuliaddress]{College of Software Engineering, Zhengzhou University of Light Industry, China}
\address[zkyaddress]{Institute of Automation, Chinese Academy of Sciences, China}
\address[ruidianaddress3]{School of Information Technology, Halmstad University, Sweden}
\address[polyuaddress]{The Hong Kong Polytechnic University, HongKong}

\begin{abstract}
In the era of large language models (LLMs), the task of ``System I''~-~the fast, unconscious, and intuitive tasks, e.g., sentiment analysis, text classification, etc., have been argued to be successfully solved. However, sarcasm, as a subtle linguistic phenomenon, often employs rhetorical devices like hyperbole and figuration to convey true sentiments and intentions, involving a higher level of abstraction than sentiment analysis. There is growing concern that the argument about LLMs' success may not be fully tenable when considering sarcasm understanding. 
To address this question, we select eleven SOTA LLMs and eight SOTA pre-trained language models (PLMs) and present comprehensive evaluations on six widely used benchmark datasets through different prompting approaches, i.e., zero-shot input/output (IO) prompting, few-shot IO prompting, chain of thought (CoT) prompting. We build SarcasmBench, the first benchmark to comprehensively evaluate LLMs on sarcasm understanding. Our results highlight three key findings:
(1) current LLMs underperform supervised PLMs based sarcasm detection baselines. This suggests that significant efforts are still required to improve LLMs' understanding of human sarcasm.
(2) GPT-4 consistently and significantly outperforms other LLMs across various prompting methods, with an average improvement of 14.0\%$\uparrow$. 
(3) Few-shot IO prompting method outperforms the other two methods: zero-shot IO and few-shot CoT. 
We hope this paper can provide some guidance for subsequent researchers to better solve sarcasm detection in the era of LLMs.
\end{abstract}


\end{frontmatter}


\section{Introduction}\label{section:introduction}
Recent large language models (LLMs) have demonstrated outstanding instruction-following and in-context learning abilities across various natural language processing (NLP) tasks, such as question answering~\cite{shao2023prompting}, sentiment analysis~\cite{zhang2023dialoguellm}, text classification~\cite{zhang2024pushing}, etc. In the era of LLMs, it has been argued that ``System I'' tasks~-~the fast, unconscious, and intuitive tasks have been successfully solved~\cite{wei2022chain}. Persistent focus and efforts from both the academic and industrial sectors have primarily concentrated on what is known as ``System II'' tasks. Such tasks demand slow, deliberate, and multiple-step cognitive processes, including logical, mathematical, and commonsense reasoning~\cite{yao2024tree}.

 \begin{figure}[t]
    \centering
    \includegraphics[width=3in]{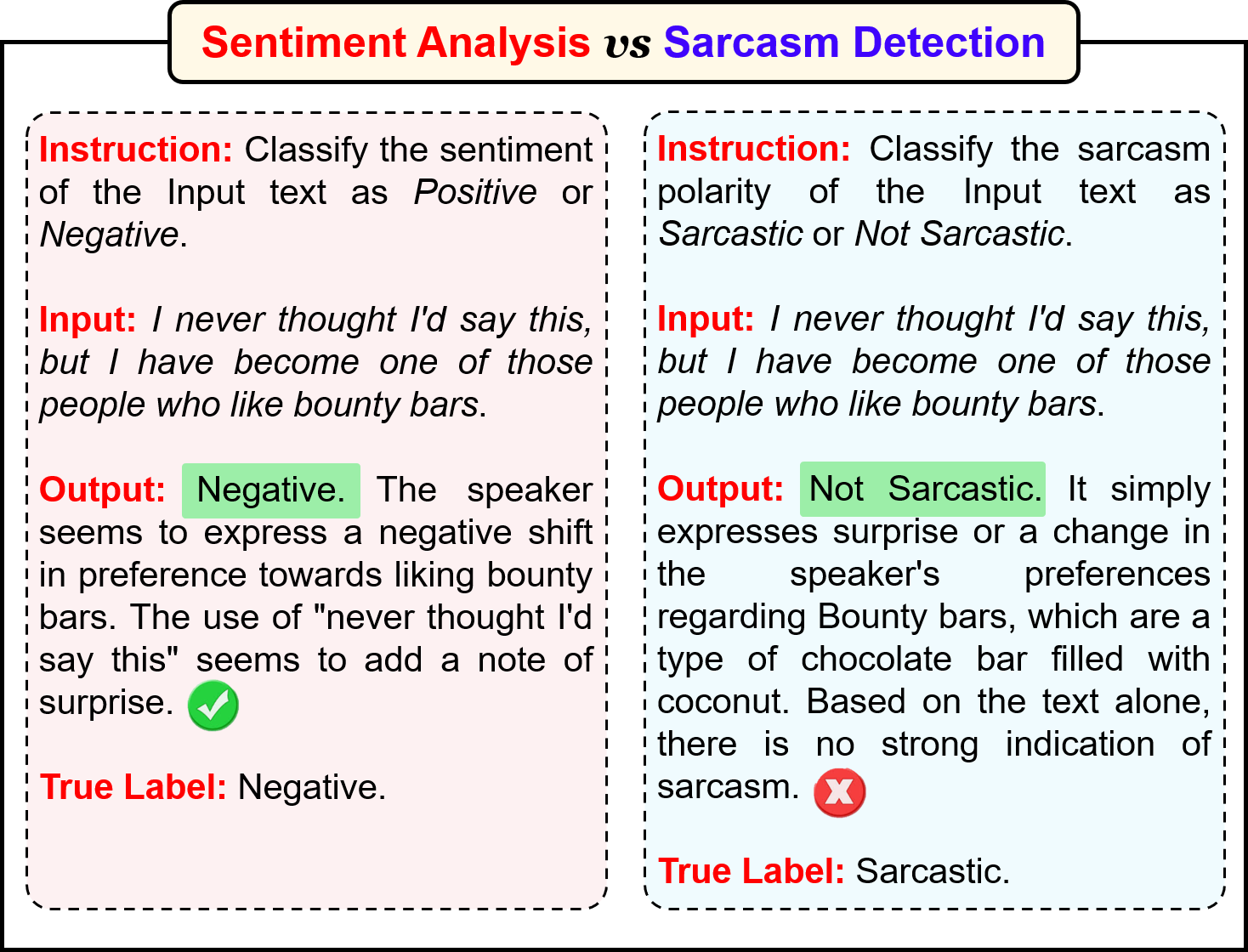}
  \caption{The comparison of sentiment analysis and sarcasm detection via a LLM.}
   \label{fig:example}
\end{figure}
Sarcasm is a subtle linguistic phenomenon that employs rhetorical devices like hyperbole and figuration to convey true sentiments and intentions that are opposite to the literal meanings of the words used~\cite{ren2023knowledge}. Humans might say something that sounds positive on the surface, but in reality, they are expressing a negative sentiment. For example, the sentence ``I like to be reprimanded." appears to convey a positive sentiment because it includes the word ``like", which typically signifies positive emotion. However, ``reprimanded" generally implies criticism and negative feedback. Sarcasm detection aims to determine whether a given text is sarcastic or non-sarcastic by leveraging different types of information, such as linguistic features, contextual information, etc~\cite{zhang2023learning}.
Due to its inherently ambiguous and metaphorical nature, sarcasm detection has consistently presented significant challenges, evolving from the era of feature engineering to that of prompt engineering~\cite{yue2023knowlenet,zhang2023stance}. 

There is growing concern that the argument about LLMs' success may not be fully tenable when considering sarcasm understanding, as it involves a higher level of abstraction than sentiment analysis, as shown in Fig.~\ref{fig:example}. Hence, the main research question can be written as:

\textbf{RQ:}  \textit{Have LLMs really made significant progress in understanding sarcasm?}

To answer this question, we propose an evaluation framework that consists of various combinations of LLMs and prompting approaches, systematically assessing the performance of LLMs on sarcasm detection. In particular, we select eleven SOTA LLMs, including ChatGPT\footnote{https://chat.openai.com/}, GPT-4~\cite{achiam2023gpt}, Claude 3\footnote{https://claude.ai/}, Mistral~\cite{jiang2023mistral}, Baichuan 2/3~\cite{yang2023baichuan}, ChatGLM 2/3~\cite{glm2024chatglm}, LLaMA 2/3~\cite{touvron2023llama}. Qwen 1.5~\cite{yang2024qwen2}, and eight PLMs to evaluate their performance on six benchmarks using three popular prompting methods: zero-shot IO prompting, few-shot IO prompting, and CoT prompting.

The experimental results highlight three key findings:
(1) Current prompting-based LLMs underperform supervised PLMs across six benchmark datasets. This suggests that significant efforts are still required to improve LLMs' understanding of human sarcasm.
(2) GPT-4 consistently and significantly outperforms other LLMs across various prompting methods, with an average improvement of 14.0\%$\uparrow$. Claude 3 and ChatGPT demonstrate the next best performance after GPT-4.
(3) Few-shot IO prompting method outperforms the other two methods: zero-shot IO and few-shot CoT, with an average improvement of 4.5\%$\uparrow$. The reason is that sarcasm detection, being a holistic, intuitive, and non-rational cognitive process, is argued not to adhere to step-by-step logical reasoning, making CoT less effective in understanding sarcasm compared to its effectiveness in mathematical reasoning tasks. This also aligns closely with the key findings of the SarcasmCue framework~\cite{yao2024sarcasmdetectionstepbystepreasoning}.
These findings underscore the complex nature of sarcasm detection and the current limitations of LLMs, while also highlighting the further need of stronger LLMs.

The main contributions are concluded as follows:
\begin{itemize}
\item This is the first work that comprehensively evaluates the performance of LLMs in understanding sarcasm.

\item We propose an evaluation framework that consists of various combinations of eleven LLMs and three prompting approaches.

\item Comprehensive experiments over six datasets demonstrate the strengths and the limitations of current LLMs.
\end{itemize}

\section{Related Work}\label{relatedwork}
This section reviews two lines of research that form the basis of this work: sarcasm detection and large language models.

\subsection{Sarcasm Detection}
Sarcasm detection aims to identify whether the given text is sarcastic or not~\cite{zhang2024cmma}. It has evolved from early rule based and statistical learning based approaches to traditional neural methods, such as CNN, RNN, and further advanced to modern neural methods epitomized by Transformer models. In the early stage, the rule based approaches infer the overall sarcasm polarity based on the refined sarcasm rules, such as the occurrence of the interjection word~\cite{zhang2023stance}. Statistical learning based approaches mainly employ statistical learning techniques, e.g., SVM, RF, NB, etc., to extract patterns and relationships within the data~\cite{zhou2023bns}.

As deep learning has shown the superiority over statistical learning, numerous base neural networks, e.g., such as CNN~\cite{jain2020sarcasm}, LSTM~\cite{ghosh2018sarcasm}, GCN~\cite{liang2022multi}, etc., have been predominantly utilized during the middle stage of sarcasm detection research, aiming to learn and extract complex features in an end-to-end fashion. As the field of deep learning continues to evolve, sarcasm detection research has stepped into the era of pre-trained language models (PLMs). An increasing number of researchers are designing sophisticated PLM architectures to serve as encoders for obtaining effective text representations. For example, Liu et al.~\cite{liuetal2022dual} proposed a dual-channel framework by modeling both literal and implied sentiments separately~\cite{liuetal2022dual}. They also constructed two conflict prompts to elicit PLMs to generate the sarcasm polarity~\cite{liu2023prompt}. Qiao et al. presented a mutual-enhanced incongruity learning network to take advantage of the underlying consistency between the two modules to boost the performance~\cite{qiao2023mutual}. Tian et al. proposed a dynamic routing Transformer network to activate different routing transformer modules for modeling the dynamic mechanism in sarcasm detection~\cite{tianetal2023dynamic}. 

However, the above-mentioned works still focus on how to utilize PLMs to extract effective features, without leveraging the extraordinary understanding capabilities of LLMs. In contrast, this paper employs three prompting methods to make the first attempt to explore the potential of prompting LLMs in sarcasm detection.

\subsection{Large Language Models}
LLMs are advanced language models characterized by their substantial parameter sizes and exceptional learning capabilities ~\cite{chen2021evaluating}. In recent years, the community of natural language processing (NLP)  has witnessed substantial progress largely due to the development of LLMs. These models excel in areas such as in-context learning, few-shot prompting, and following complex instructions. OpenAI has been at the forefront of this innovation, notably with the introduction of transformative models such as ChatGPT and GPT-4. Nevertheless, the exclusive nature of these technologies has led to the emergence of various LLM versions, which often incorporate tens or even hundreds of billions of parameters~\cite{zhao2023survey}. We categorize these LLMs into two groups based on their specialization: general LLMs and specialized LLMs. 

General LLMs are designed for versatility across a wide spectrum of NLP tasks. Prominent examples of these models are GPT-4, ChatGLM 4, LLaMA 3\footnote{https://llama.meta.com/}, PanGu-$\Sigma$~\cite{ren2023pangu}, Baichuan 3\footnote{https://www.baichuan-ai.com}, etc. Such LLMs often perform well across a range of tasks, but their potentials in specific domains await further exploration. In contrast, specialized LLMs are fine-tuned for specific tasks via task-specific architectures and knowledge, allowing them to achieve higher performance. For example, Zhang et al. proposed a fine-tuned context and emotion knowledge tuned LLM for emotion recognition in conversations~\cite{zhang2023dialoguellm}. They also presented RGPT, an adaptive boosting framework tailored to produce a specialized text classification LLM by recurrently ensembling a pool of strong base learners~\cite{zhang2024pushing,yao2024sarcasm}.

However, the above-mentioned studies do not involve sarcasm understanding, we propose an evaluation framework to unlock its potential in sarcasm understanding.

\section{The Proposed Approach}
\subsection{Task Definition}
Sarcasm detection is transformed as a conditional generative task, where the output $\mathcal{Y}$ will be the labels. Given a set of input texts $\mathcal{X} = \{x_1, x_2, \ldots, x_{N}\}$ where each document $x_i$ is augmented with a designed prompt ${Prompt}_i \in \mathcal{P}$ that provides contextual guidance, i.e., ${Prompt}_i =  {INS}_i \oplus x_i  $, where ${INS}_i$ represents the task instruction, $\mathcal{P}$ represents the prompt set. 
Our task is to let a LLM $\mathcal{M}$ map an input document to its target label: $\mathcal{M}(\mathcal{X}, \mathcal{P}, \theta) \rightarrow \mathcal{Y}$, where $\mathcal{Y} = \{y_1, y_2, \ldots, y_{N}\}$ denotes the label sequence generated by the LLM $\mathcal{M}$. We formulate the problem as:
\begin{equation} 
\begin{aligned}
\mathcal{M} =\mathrm{arg}~\underset{c}{max}\prod_{i}Prob\left ( y_i=c|x_i, {INS}_i, \theta \right ) 
\end{aligned}
\end{equation}

 \begin{figure*}[t]
    \centering
    \includegraphics[width=5.5in]{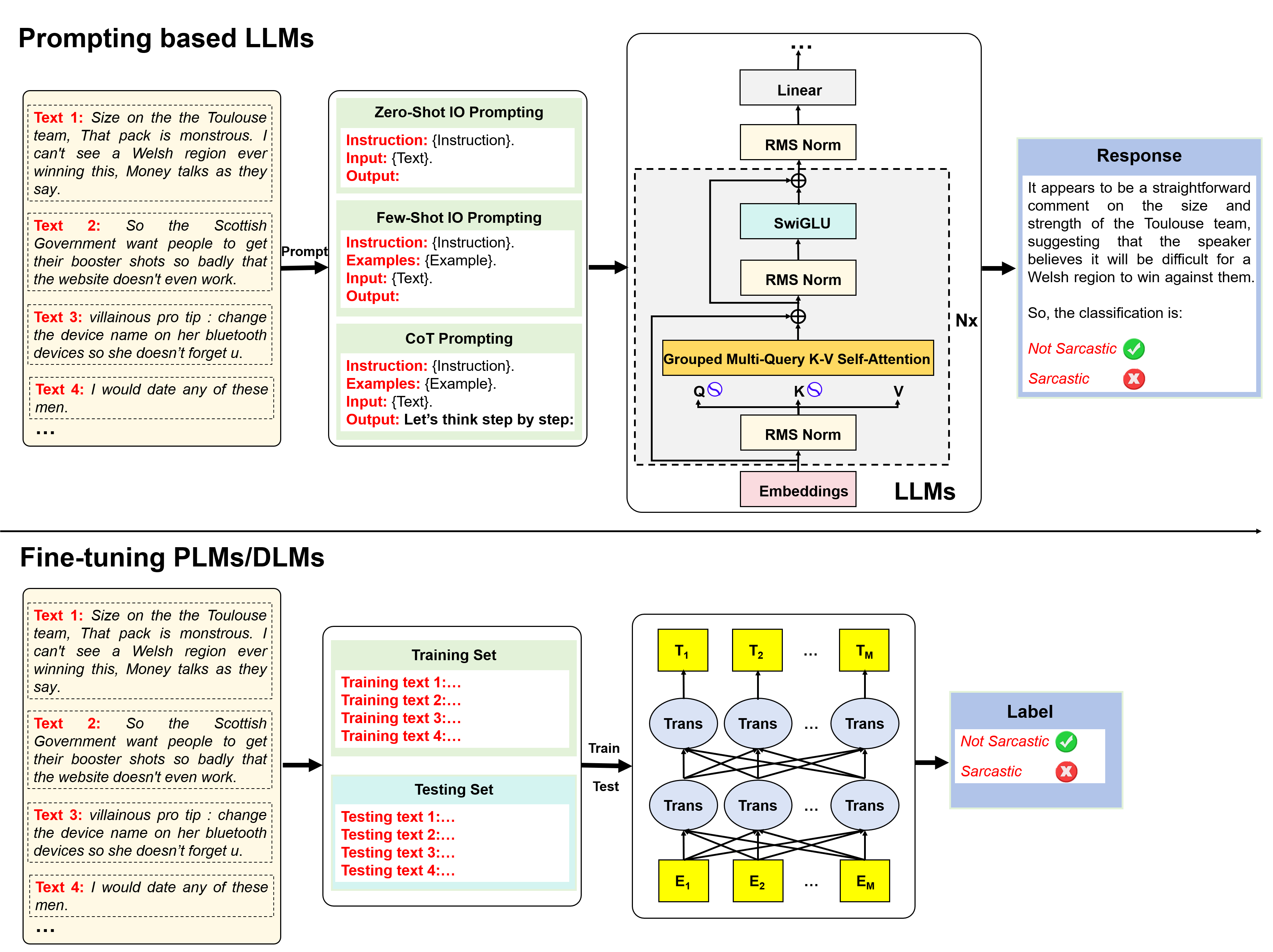}
  \caption{The proposed evaluation framework for LLMs and PLMs/DLMs.}
   \label{fig:model}
\end{figure*}

\subsection{Evaluation Approach}
Fig.~\ref{fig:model} shows the overview of our proposed evaluation framework. 
Initially, we take both sarcastic and non-sarcastic samples as inputs. Subsequently, we design prompts tailored specifically for sarcasm. This involves constructing task instructions and providing examples for few-shot learning. Our evaluation utilizes three distinct prompting approaches: zero-shot IO prompting, few-shot IO prompting, and CoT prompting. These methods are implemented to assess how effectively LLMs generate responses that are both valid and rigorously aligned with the subtleties of sarcastic expressions. Finally, we analyze and compare the outputs generated by the LLMs against the ground truth to evaluate the models' proficiency in understanding and detecting sarcasm.

\subsection{Prompt Construction}
Our prompt ${Prompt}_i$ contains of three key components:

\textbf{(1) Task instruction.}  It clearly defines the specific sarcasm detection task the model needs to accomplish. In this work, the task instruction is given as follows:

\textit{This is a sarcasm classification task. Determine whether the following input text expresses sarcasm, if it does, output `sarcastic', otherwise, output `non-sarcastic'. Return the label only without any other text.}

\textbf{(2) Input.}  The input is every testing sample $x_i$ to classify.

\textbf{(3) Demonstration.}  It provides specific examples of how to analyze text for sarcasm, showing both sarcastic and non-sarcastic responses to similar situations. It will guide the LLMs in recognizing subtle cues and patterns that identify sarcasm. In addition, it also provides an output format that LLM’s outputs should follow. Note that demonstrations are only needed for the few-shot setup, but not for the zero-shot setup. 

We adopt KNN search~\cite{sun2023text} to sample examples that are similar to the test sequence. In this method, the text initially undergoes transformation into a vector via the encoding function $f$. This vector is then employed as a query to traverse the complete training dataset, aiming to identify the $k$ text sequences closest to it. From these, we extract the $k$ most similar data examples to use as demonstrations.

\subsection{Three Prompting Approaches}
\textbf{Zero-shot IO Prompting} refers to the scenario where LLMs generate the output $y$ based purely on the input $x$, without any additional examples or guidance. Mathematically, this can be represented as:
\[
y \sim p_{\theta}(y \mid x),
\]
where $p_{\theta}$ represents the pre-trained model, and $x$ is directly passed to the model to generate $y$. The model must rely solely on its prior knowledge and understanding of the task.

\textbf{Few-shot IO Prompting} involves providing LLMs with a limited number of examples or demonstrations of the task before predicting the output for a new input $x$. This can be formalized as:
\[
y \sim p_{\theta}(y \mid \text{prompt}_{\text{IO}}(x)),
\]
where $\text{prompt}_{\text{IO}}(x)$ embeds input $x$ within a context of task instructions and few-shot examples. In this approach, $p_{\theta}$ adapts based on the provided examples, which helps guide the model in generating more accurate outputs.

\textbf{Few-shot CoT prompting} extends the idea of few-shot prompting by explicitly modeling intermediate reasoning steps between the input $x$ and the final output $y$. The LLM is guided through a sequence of logical steps $z_1, z_2, \dots, z_n$, leading to the final answer. This process is described as:
\[
[z_1, z_2, \dots, z_n, y] \sim p_{\text{CoT}\, \theta}(z_1, z_2, \dots, z_n, y \mid x),
\]
where each $z_i$ represents a coherent reasoning step. By incorporating these intermediate steps, the model is better able to handle complex tasks, such as solving multi-step problems or answering detailed questions.

\section{Experiments}
\subsection{Dataset}
\textbf{Datasets.} Six benchmark datasets are selected as the experimental beds which encompass the most significant datasets used in sarcasm detection tasks. They are IAC-V1~\cite{lukin2017really}, IAC-V2~\cite{oraby2017creating}, Ghosh~\cite{ghosh2016fracking}, iSarcasmEval~\cite{farha2022semeval}, Riloff~\cite{riloff2013sarcasm} and SemEval 2018 Task 3~\cite{van2018semeval}. 

\textbf{IAC-V1} and \textbf{IAC-V2} are from the Internet Argument Corpus (IAC)~\cite{lukin-walker-2013-really}, specifically designed for the task of identifying and analyzing sarcastic remarks within online debates and discussions. It encompasses a balanced mixture of sarcastic and non-sarcastic comments.

\textbf{Ghosh} consists of 51,189 tweets (24,453 sarcastic tweets and 26,736 non-sarcastic tweets) in which sarcastic tweets are automatically collected from Twitter using the user's self-declaration of sarcasm/irony with sarcastic and ironic hashtags (e.g. \#irony, \#sarcasm). In this work, we have conducted a thorough double-check of this dataset and successfully filtered out 7,804 noisy tweets.

\textbf{iSarcasmEval} is the first shared task to target intended sarcasm detection. Each sample in this dataset is provided and labelled by the authors of the texts themselves. For sarcastic texts, there is a rephrase that conveys the same message non-sarcastically. For English sarcastic texts, there is a label specifying the category of ironic speech that it reflects.

\textbf{Riloff} collects a set of 1,600 tweets that contain \#sarcasm or \#sarcastic, and another 1,600 without these tags. It chooses to remove such tags from all tweets and present the tweets to a group of human annotators for final labelling.

\textbf{SemEval 2018 Task 3} is collected using irony-related hashtags (i.e. \#irony, \#sarcasm, \#not) and are subsequently manually annotated to minimize the amount of noise in the corpora. It emhpasizes the challenges inherent in identifying sarcasm within the constraints of MUStARD's concise format and highlights the importance of context and linguistic subtleties in recognizing sarcasm.
The statistics for each dataset are shown in Table~\ref{tab:statistics}.

\subsection{Evaluation Metrics}
To make a fair comparison, we employ \textit{precision} (P), \textit{recall} (R), \textit{accuracy} (Acc) and  \textit{F1}  as evaluation metrics. 
For each method, we run five random seeds and report the average result of the test sets. The metrics are written as:
\begin{equation}
 \begin{aligned}
 \label{eq}
     &P=\frac{TP}{TP+FP},  \\
    &R=\frac{TP}{TP+FN},  \\
    &F1=\frac{2\ast P\ast R}{P+R}, \\
    &Acc=\frac{TP+TN}{TP+FN+FP+TN}, 
 \end{aligned} 
\end{equation}
where TP (True Positives) is the number of positive samples correctly identified, FP (False Positives) is the number of negative samples incorrectly labeled as positive, FN (False Negatives) is the number of positive samples incorrectly labeled as negative,
TN (True Negatives) is the number of negative samples correctly identified.
\begin{table}[t]
\caption{Dataset statistics.} \label{tab:statistics}
\scalebox{0.78}{
\begin{tabular}{lccccclll}
\cline{1-6}
\textbf{Dataset}         & \textbf{Train} & \textbf{Valid} & \textbf{Test} & \textbf{Sarcastic} & \textbf{Non-sarcastic} &  &  &  \\ \cline{1-6}
IAC-V1          & 1214                      & 304                       & 417                      & 973                           & 962                               &  &  &  \\
IAC-V2          & 4031                      & 1008                      & 1481                     & 3260                          & 3260                              &  &  &  \\
Ghosh           & 32708                     & 8687                      & 2000                     & 19546                         & 23839                             &  &  &  \\
iSarcasmEval    & 821                       & 280                       & 299                      & 200                           & 1200                              &  &  &  \\
Riloff          & 276                       & 100                       & 113                      & 77                            & 412                               &  &  &  \\
SemEval Task 3 & 2910                      & 924                       & 784                      & 2222                          & 2396                              &  &  &  \\ \cline{1-6}
                & \multicolumn{1}{l}{}      & \multicolumn{1}{l}{}      & \multicolumn{1}{l}{}     & \multicolumn{1}{l}{}          & \multicolumn{1}{l}{}              &  &  &  \\
                & \multicolumn{1}{l}{}      & \multicolumn{1}{l}{}      & \multicolumn{1}{l}{}     & \multicolumn{1}{l}{}          & \multicolumn{1}{l}{}              &  &  & 
\end{tabular}
}
\end{table}

\subsection{Implementation Details} 
We use the official implementations and hyper-parameters for all traditional deep learning models (DLMs) and PLMs based approaches. For example, all weight matrices are given their initial values by sampling from a uniform distribution $U(-0.1, 0.1)$. The learning rate is set to 5e-5. The batch size is set to 32 and the number of epochs is set to 30. The dropout rate is set to 0.5. In contrast, the LLMs based methods are implemented with the respective official Python API library and the Hugging Face Transformers library\footnote{https://huggingface.co/docs/transformers}. 
All the experiments are conducted on an NVIDIA GeForce GTX 4090 GPU. 

\subsection{Compared Baselines}
A wide range of strong baselines are included for comparison including DLMs, PLMs and LLMs. They are:

\begin{itemize}
\item \textit{\textbf{Random:}}

\textbf{(1) Random} This baseline makes random predictions sampled uniformly across the test set.

\item \textit{\textbf{DLMs based approaches:}}

\textbf{(2) TextCNN~\cite{kim2014convolutional}} designs a CNN~\cite{kim2014convolutional} with three convolutional layers and a fully connected layer. It is trained on top of word embeddings for utterance-level classification.

 \textbf{(3) LSTM} takes word embeddings as input and takes the hidden representation of each word for sarcasm classification.

  \textbf{(4) Bi-LSTM~\cite{ijcai2019p755}} implements a bidirectional LSTM to learn both forward and backward long-range dependency information.

\textbf{(5) AT-LSTM~\cite{wang2016attention}:} is an attention-based LSTM with aspect embedding. We obtain aspect embeddings by averaging the vectors of words, and append it with each word embedding vector.
 \begin{figure*}[t]
\centering
\includegraphics[width=5.5in]{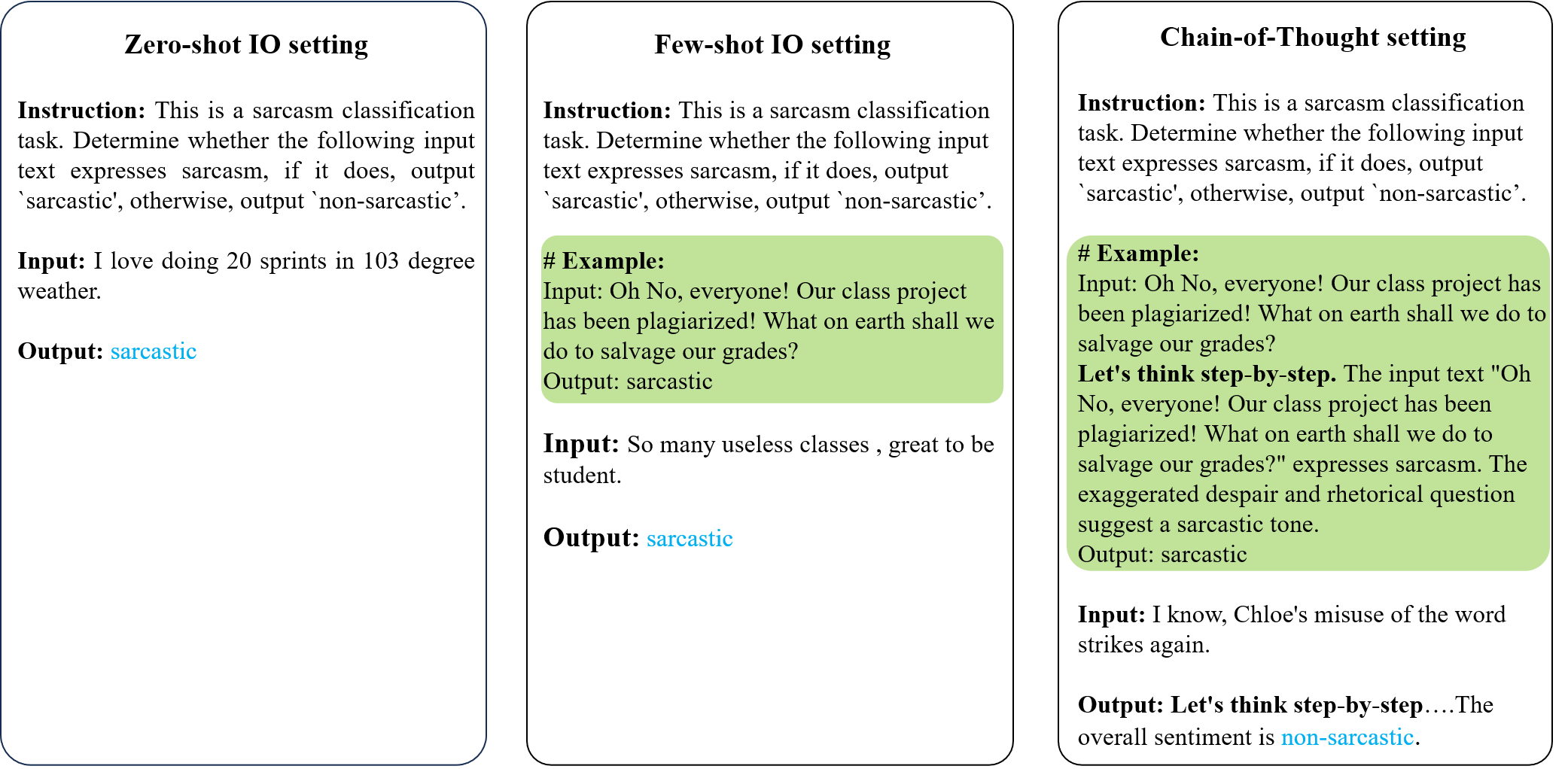}
\caption{Examples of three prompting approaches.} \label{fig:promptingapp}
\end{figure*}
 
 \item \textit{\textbf{PLMs based approaches:}}

  \textbf{(6) BERT~\cite{devlin-etal-2019-bert}} produces contextualized word embeddings for sarcasm detection.
  
 \textbf{(7) RoBERTa~\cite{liu2019robertarobustlyoptimizedbert}} builds on BERT and modifies key hyperparameters, removing the next-sentence pretraining objective and training with much larger mini-batches and learning rates.

 \textbf{(8) DeBERT~\cite{he2021deberta}} improves the BERT and RoBERTa models using the disentangled attention mechanism. 
 
  \textbf{(9) XLNet~\cite{yang2019xlnet}} improves the BERT through a generalized autoregressive pretraining method enables learning bidirectional contexts by maximizing the expected likelihood over all permutations of the factorization order. 

 \textbf{(10) DC-Net-RoBERTa~\cite{liu-etal-2022-dual}} designs the dual-channel network (DC-Net) to recognize sentiment conflict by modeling both literal and implied sentiments separately.

\item \textit{\textbf{LLMs based approaches:}}

\textbf{(11) Baichuan 2~\cite{yang2023baichuan}} is a Chinese LLM trained from scratch, on 2.6 trillion tokens. It matches or outperforms other open-source models of similar size on public benchmarks.  

\textbf{(12-13) ChatGLM 2/3~\cite{glm2024chatglm}} are two open bilingual language model based on General Language Model (GLM) framework, with 6.2 billion parameters. 

  \textbf{(14-15) LLaMA 2/3~\cite{touvron2023llama}} are trained on 2 trillion tokens, and have double the context length than Llama 1, and outperform other open source language models on many external benchmarks.
 
  \textbf{(16) Mistral~\cite{jiang2023mistral}} consistently outperforms Llama2-13B on all metrics and stands competitively with Llama-34B.

  \textbf{(17-18) Qwen 1.5/2~\cite{yang2024qwen2}} are two SOTA LLMs trained on data in 27 additional languages besides English and Chinese and achieves state-of-the-art performance in a large number of benchmark evaluations.

\textbf{(19-20) GPT-3.5/4~\cite{achiam2023gpt}} are considered as two strongest general LLM.

\textbf{(21) Claude 3\footnote{https://claude.ai/}} is Anthropic's second-most powerful AI model, with strong performance on highly complex tasks.

\end{itemize}

\begin{table*}[t!]
\centering
\caption{Performance on six datasets. For LLMs, all strategies are based on a \textbf{zero-shot IO setting}. \textbf{\textcolor{blue}{Blue}} indicates the best results across LLMs.}
\label{tab:baseline1}
\scalebox{0.82}{
\begin{tabular}{cl|cccc|cccc|cccc|c}
\midrule[1pt]
                                                   &                         & \multicolumn{4}{c}{\textbf{IAC-V1}}                                     & \multicolumn{4}{c}{\textbf{IAC-V2}}                                     & \multicolumn{4}{c}{\textbf{iSarcasmEval}}                               &                                       \\ \cline{3-14}
                                                   
\multirow{-2}{*}{\textbf{Paradigm}}  & \multirow{-2}{*}{\textbf{Model}} & \textbf{Acc}          & \textbf{P}          & \textbf{R}           & \textbf{F1}            & \textbf{Acc}          & \textbf{P}          & \textbf{R}           & \textbf{F1}   & \textbf{Acc}          & \textbf{P}          & \textbf{R}           & \textbf{F1}           & \multirow{-2}{*}{\textbf{Avg. of F1}}           \\\midrule[1pt]

 Random  & Random                    &52.0             & 62.5          & 48.4          & 54.5          & 51.9   & 52.1          & 53.2          & 52.7          & 56.5             &    15.2         &   52.6 &    23.5       & \cellcolor[HTML]{E5FBBB}43.5         \\
 
 \hline

                                                        & LSTM                    & 55.3             & 54.9          & 54.9           & 54.8          & 68.5             & 72.3          & 68.0           & 68.8          & 72.7             &    57.0          &   55.1          &    55.6         & \cellcolor[HTML]{E5FBBB}59.7         \\

                                                       & TextCNN                 & 54.2             & 53.7             & 53.6              & 53.6             & 68.2             & 68.3             & 67.7              & 67.9             & 72.1             & 54.5             & 53.1              & 53.4             & \cellcolor[HTML]{E5FBBB}58.3             \\
                           
       & Bi-LSTM                 & 64.6          & 64.6          & 64.6           & 64.6          & 79.7          & 79.8          & 79.7           & 79.7          & 74.9          & 57.0          & 55.4           & 55.8          & \cellcolor[HTML]{E5FBBB}66.7          \\ 

 \multirow{-4}{*}{DLMs}   & AT-LSTM                    &65.5            & 65.9          & 65.5 & 65.3          &  76.2 & 76.7          & 76.2          & 76.1          & 73.5             &    55.8         & 53.2          &    54.6       & \cellcolor[HTML]{E5FBBB}65.3         \\

\hline

                             & BERT                    & 65.3             & 65.4             & 65.1 & 65.2             & 76.4             & 76.5             & 76.1              & 76.2             & 74.0 & 62.5          & 58.2           & 58.8          & \cellcolor[HTML]{E5FBBB}66.8             \\

                            & RoBERTa                 & \textbf{\underline{70.1}}          & \textbf{\underline{70.0}}          & \textbf{\underline{70.1}}           & \textbf{\underline{69.9}}          & 80.7          & 80.9          & 80.8           & 80.7          & 78.9          & 66.4          & 57.5           & 63.5          & \cellcolor[HTML]{E5FBBB}71.3          \\

                                                     & DeBERT                   & 66.7             & 66.8             & 66.7              & 66.6             & 78.3             & 78.4             & 78.1              & 78.2   & 71.2 & 58.4             & 55.6              & 56.9            & \cellcolor[HTML]{E5FBBB}67.2             \\

\multirow{-4}{*}{PLMs}             & DC-Net-RoBERTa          & 69.3          & 69.7          & 69.3           & 69.1          & \textbf{\underline{81.7}}          & \textbf{\underline{81.7}}          & \textbf{\underline{81.7}}           & \textbf{\underline{81.7}}          & \textbf{\underline{79.5}}             & \textbf{\underline{67.1}}             & \textbf{\underline{58.3}}              & \textbf{\underline{64.0}}             & \cellcolor[HTML]{E5FBBB}\textbf{\underline{71.5}}          \\
 \hline
 
                                                      & Baichuan 2-7B            & 57.3          & 67.5          & 54.4           & 60.2          & 62.8          & 62.3          & 65.9           & 64.0          & 38.5          & 12.4          & 63.2           & 20.7          & \cellcolor[HTML]{E5FBBB}48.3          \\

                                                & ChatGLM 2-6B            & 47.0          & 69.0          & 19.8           & 30.7          & 51.0          & 54.5          & 15.6           & 24.2          & \textcolor{blue}{\textbf{\underline{66.6}}} & 6.9           & 13.2           & 9.1           & \cellcolor[HTML]{E5FBBB}21.3          \\
                           
                                                       & ChatGLM 3-6B             & 54.0          & 62.1          & 58.1           & 60.0          & 61.8          & 59.8          & 73.3           & 65.9          & 66.6          & 20.2          & 55.3           & 30.0          & \cellcolor[HTML]{E5FBBB}52.0          \\
                           
                                                     & LLaMA 2-7B               & 59.2          & 59.7          & 96.8           & 73.9          & 50.6          & 50.4          & 97.3           & 66.4          & 19.1          & 13.3          & 97.4           & 23.4          & \cellcolor[HTML]{E5FBBB}54.6          \\

                            & LLaMA 3-8B               & 60.4          & 60.1          & 93.5           & 73.8          & 52.1          & 51.2          & 97.8           & 67.2          & 13.4          & 12.8          &\textcolor{blue}{\textbf{\underline{100.0}}}           & 22.7          & \cellcolor[HTML]{E5FBBB}54.6          \\

                                                    & Mistral-7B              & 61.9          & 67.6          & 69.0           & 68.3          & 57.9          & 56.1          & 75.0           & 64.2          & 46.8          & 16.2          & 76.3           & 26.7          & \cellcolor[HTML]{E5FBBB}53.1          \\

                                                       & Qwen 1.5-7B            & 59.2          & 59.4          & \textcolor{blue}{\textbf{\underline{99.6}}}  & 74.4          & 50.0          & 50.1          & \textcolor{blue}{\textbf{\underline{99.5}}}  & 66.7          & 12.7          & 12.7          & \textcolor{blue}{\textbf{\underline{100.0}}} & 22.6          & \cellcolor[HTML]{E5FBBB}54.6          \\
                                                       
                                                            & Qwen 2-7B            & 56.6          &36.3           & 76.9          &49.3         & 51.8          &28.9        & 57.8          &38.6       & 46.1 &18.2 & 36.4          &24.3          & \cellcolor[HTML]{E5FBBB}37.4          \\

                                                                 & ChatGPT            & 63.6          & 61.2          & 81.8          &70.0         & 56.4         &50.2       & 91.6          &64.9      & 51.5  &14.3         & 91.7          &26.2          & \cellcolor[HTML]{E5FBBB}53.7          \\

                                                       & GPT-4 Turbo         & \textbf{\textcolor{blue}{\underline{72.7}}} & \textcolor{blue}{\textbf{\underline{73.3}}} & 85.1           & \textcolor{blue}{\textbf{\underline{78.7}}} & \textcolor{blue}{\textbf{\underline{71.4}}} & \textcolor{blue}{\textbf{\underline{65.1}}} & 92.9           & \textcolor{blue}{\textbf{\underline{76.6}}} & 65.6          & \textcolor{blue}{\textbf{\underline{25.6}}} & 89.5           & \textcolor{blue}{\textbf{\underline{39.8}}} & \cellcolor[HTML]{E5FBBB}\textcolor{blue}{\textbf{\underline{65.0}}} \\
                                                       
\multirow{-11}{*}{LLMs$_{+\text{\textbf{0-shot~IO}}}$}    & Claude-3-haiku          & 66.7          & 65.3          & 94.0           & 77.0          & 58.3          & 54.7          & 97.8           & 70.2          & 31.4          & 15.4          & 97.4           & 26.5          & \cellcolor[HTML]{E5FBBB}57.9          \\ \midrule[1pt]

                                                     &                         & \multicolumn{4}{c}{\textbf{Riloff}}                                     & \multicolumn{4}{c}{\textbf{SemEval Task 3}}                            & \multicolumn{4}{c}{\textbf{Ghosh}} &                                       \\ \cline{3-14}
                                                     
\multirow{-2}{*}{\textbf{Paradigm}}  & \multirow{-2}{*}{\textbf{Model}} & \textbf{Acc}          & \textbf{P}          & \textbf{R}           & \textbf{F1}            & \textbf{Acc}          & \textbf{P}          & \textbf{R}           & \textbf{F1}   & \textbf{Acc}          & \textbf{P}          & \textbf{R}           & \textbf{F1}           & \multirow{-2}{*}{\textbf{Avg. of F1}}           \\\midrule[1pt]

Random & Random & 50.4 & 5.8 & 30.0 & 9.7 & 48.9 & 38.5 & 48.2 & 42.8 & 49.8 & 49.8 & 51.4 & 50.6 &   \cellcolor[HTML]{E5FBBB}34.3 \\\hline 

                                                     & LSTM                    & 76.0      & 61.0 & 64.0       & 62.0      & 73.2      & 65.4      & 68.7       & 67.3      & 70.4             & 73.2             & 58.7              & 65.2             & \cellcolor[HTML]{E5FBBB}65.0             \\
                                                     
                                                      & TextCNN                 & 75.1      & 62.8      & 62.2      & 62.7      & 64.0      & 65.0      & 64.0       & 63.0      &\textbf{\underline{84.8}}            & \textbf{\underline{83.3}} & \textbf{\underline{82.9}}              & \textbf{\underline{83.1}}             & \cellcolor[HTML]{E5FBBB}69.6             \\

             & Bi-LSTM                 & 69.6      & 60.5      & 64.2      & 61.7              & 66.4             & 60.5             & 66.7              & 63.4            & 74.7             & 70.8             & 67.9              & 69.3             & \cellcolor[HTML]{E5FBBB}64.8             \\

\multirow{-4}{*}{DLMs}             & AT-LSTM                 & 77.7             & 58.5             & 60.4              & 59.6             & 72.8             & 64.8             &69.3             & 67.0             & 75.2             & 68.7             & 69.8              & 70.1             & \cellcolor[HTML]{E5FBBB}65.6             \\ 
\hline

                                                     & BERT                    & 79.7          & \textbf{\underline{72.6}}          & 56.6           & 63.6          & 69.6             & 63.5             & 74.1              & 68.4             & 79.2          & 70.9          & 80.6           & 78.8          & \cellcolor[HTML]{E5FBBB}70.2             \\

                                                    & RoBERTa                 & \textbf{\underline{84.0  }}    & 64.0      & \textbf{\underline{75.0}}      & \textbf{\underline{69.0}}     & \textbf{\underline{75.0}}      & 64.0      & \textbf{\underline{84.0}}       & \textbf{\underline{72.0}}      & 80.3             & 80.3             & 79.8              & 80.0            & \cellcolor[HTML]{E5FBBB}\textbf{\underline{73.7}}             \\

                                        & XLNet                   & 83.5      & 52.0      & 73.0      & 61.0     & 66.0     & 55.0      & 83.0       & 66.0      & 77.2            & 77.7             & 78.2   & 77.9            & \cellcolor[HTML]{E5FBBB}68.3             \\

\multirow{-4}{*}{PLMs}            & DC-Net-RoBERTa          & 82.6      & 67.4      & 70.3      & 67.8            & 70.9             & \textbf{\underline{69.7}}            & 68.3              & 68.7             & 82.2           &81.0             & 81.7              & 81.3             & \cellcolor[HTML]{E5FBBB}72.6             \\ \hline

                                                   & Baichuan 2-7B            & 25.7          & 9.8           & 90.0           & 17.6          & 47.2          & 41.9          & 85.5           & 56.2          & 55.7          & 53.7          & 82.4           & 65.0          & \cellcolor[HTML]{E5FBBB}46.3          \\

                                              & ChatGLM 2-6B            & 67.3          & 0.9           & 0.0            & 0.0           & 54.0          & 32.1          & 14.5           & 20.0          & 41.1          & 24.1          & 8.3            & 12.4          & \cellcolor[HTML]{E5FBBB}10.8          \\
                                              
                                                      & ChatGLM 3-6B             & \textbf{\textcolor{blue}{\underline{74.3}}} & 14.8          & 40.0           & 21.6          & 74.1          & \textbf{\textcolor{blue}{\underline{62.8}}} & 85.2           & 72.3          & 64.1          & 66.7          & 56.2           & 61.0          & \cellcolor[HTML]{E5FBBB}51.6          \\

                                                    & LLaMA 2-7B               & 15.0          & 9.4           & \textbf{\textcolor{blue}{\underline{100.0}}} & 17.2          & 41.2          & 39.9          & 95.8           & 56.4          & 49.7          & 49.8          & 97.7           & 66.0          & \cellcolor[HTML]{E5FBBB}46.5          \\

                            & LLaMA 3-8B               & 9.7          & 8.9           & \textbf{\textcolor{blue}{\underline{100.0}}} & 16.4          & 40.2          & 39.9          &\textbf{\textcolor{blue}{\underline{100.0}}}         & 57.0          & 50.4          & 50.2          & 99.7           & 66.8          & \cellcolor[HTML]{E5FBBB}46.7          \\

                                                   & Mistral-7B              & 57.5          & 12.0          & 60.0           & 20.0          & 54.2          & 46.3          & 95.8           & 62.4          & 59.7          & 56.3          & 86.5           & 68.2          & \cellcolor[HTML]{E5FBBB}50.2          \\
                                                   
                                                       & Qwen 1.5-7B            & 8.8           & 8.5 & \textbf{\textcolor{blue}{\underline{100.0}}} & 16.3          & 39.7          & 39.7          & \textbf{\textcolor{blue}{\underline{100.0}}} & 56.8          & 50.0          & 50.0          & \textbf{\textcolor{blue}{\underline{100.0}}} & 66.7          & \cellcolor[HTML]{E5FBBB}46.6          \\

                                                       & Qwen 2-7B            & 51.6           & \textbf{\textcolor{blue}{\underline{20.8}}} & 30.1 & 23.4 & 45.1         & 30.7          & 51.0 & 38.3          & 65.6          &    54.4       & 87.7 & 68.3         & \cellcolor[HTML]{E5FBBB}43.3          \\

                                                        & ChatGPT            & 55.0           & 15.4 & 96.7 & 26.6          & 52.2         & 48.3          & 99.7 & 65.1          & 63.3          & 58.2          & 90.4 & 71.4         & \cellcolor[HTML]{E5FBBB}54.4          \\

                                                      & GPT-4 Turbo         & 64.6          & 20.0          & \textbf{\textcolor{blue}{\underline{100.0}}} & \textbf{\textcolor{blue}{\underline{33.3}}} & \textbf{\textcolor{blue}{\underline{76.1}}} & \textbf{\textcolor{blue}{\underline{62.8}}} & 98.1           & \textbf{\textcolor{blue}{\underline{76.5}}} & \textbf{\textcolor{blue}{\underline{79.8}}} & \textbf{\textcolor{blue}{\underline{73.5}}} & 93.3           & \textbf{\textcolor{blue}{\underline{82.2}}} & \cellcolor[HTML]{E5FBBB}\textbf{\textcolor{blue}{\underline{64.0}}} \\

\multirow{-11}{*}{LLMs$_{+\text{\textbf{0-shot~IO}}}$}     & Claude-3-haiku          & 24.8          & 10.5          & \textbf{\textcolor{blue}{\underline{100.0}}} & 19.0          & 46.8          & 42.7          & 99.7           & 59.8          & 61.2          & 56.4          & 98.8           & 71.8          & \cellcolor[HTML]{E5FBBB}50.2          \\ \midrule[1pt]

\end{tabular}
}
\end{table*}

\subsection{Results and Analysis of Zero-Shot IO Prompting}
We report all of the \textbf{Accuracy}, \textbf{Precision}, \textbf{Recall} and macro \textbf{F1} scores for 21 models in Table~\ref{tab:baseline1}. 
We highlight three key observations:

\textbf{(1) In the zero-shot IO setting, PLMs demonstrate outstanding performance, significantly surpassing traditional DLMs}. For instance, RoBERTa achieves an average F1 score of 71.3 across all datasets, while DC-Net-RoBERTa reaches 71.5. In contrast, traditional models like Bi-LSTM and AT-LSTM score lower, with F1 scores of 66.7 and 65.3, respectively. PLMs outperform traditional methods by approximately 7.8\%$\uparrow$. This superiority stems from the extensive pretraining that PLMs undergo on large-scale data, enabling them to better capture contextual information and adapt to various tasks. Although traditional deep learning models perform adequately on specific tasks, their lack of large-scale pretraining support limits their generalization capabilities, resulting in weaker performance.

\textbf{(2) Despite that current LLMs have not yet fully outperformed PLMs, their potential is evident in particular tasks}. For example, GPT-4 Turbo excels in the IAC-V1 and SemEval Task 3 datasets, achieving F1 scores of 78.7 and 76.5, respectively, significantly outperforming RoBERTa's scores of 69.9 and 72.0, with the margin of 8.8\% and 4.5\%$\uparrow$. However, LLMs fall short in datasets like iSarcasmEval and Riloff, where their performance does not meet expectations. The average F1 score shows that GPT-4 Turbo's difference from RoBERTa in the IAC-V2 and Ghosh datasets is relatively small. This performance variation may be due to LLMs' advantage in handling long texts and complex semantics, but their generalization capabilities are weaker when facing domain-specific or fine-grained tasks without specialized tuning. Therefore, while LLMs show strong potential in some tasks, their consistency and generalization across different tasks still need improvement. 

\textbf{(3) Among all 11 LLMs, GPT-4 Turbo demonstrates a comprehensive lead}. It performs exceptionally well across multiple datasets, particularly in IAC-V1 and SemEval Task 3, where it achieves F1 scores of 78.7 and 76.5, significantly higher than Claude-3-haiku's 77.0 and 57.9. Additionally, in the Ghosh dataset, GPT-4 Turbo reaches an F1 score of 82.2, far surpassing other models like ChatGPT (71.4), Claude-3-haiku (71.8) and LLaMA 3-8B (66.8). 
A significant reason for GPT-4's superiority over other models is its larger parameter size, which enhances its human language comprehension and reasoning capabilities. In contrast, most other models are smaller, with around 7B/8B parameters, which limits their ability to handle complex tasks. Moreover, GPT-4's ability to adapt to various data types and text structures across different tasks further contributes to its outstanding performance.
Thus, GPT-4's strong results underscore its leading position among LLMs and its powerful generalization capabilities.

\begin{figure}[t!]
     \centering
    \subfloat[]{\includegraphics[width=2.6in]{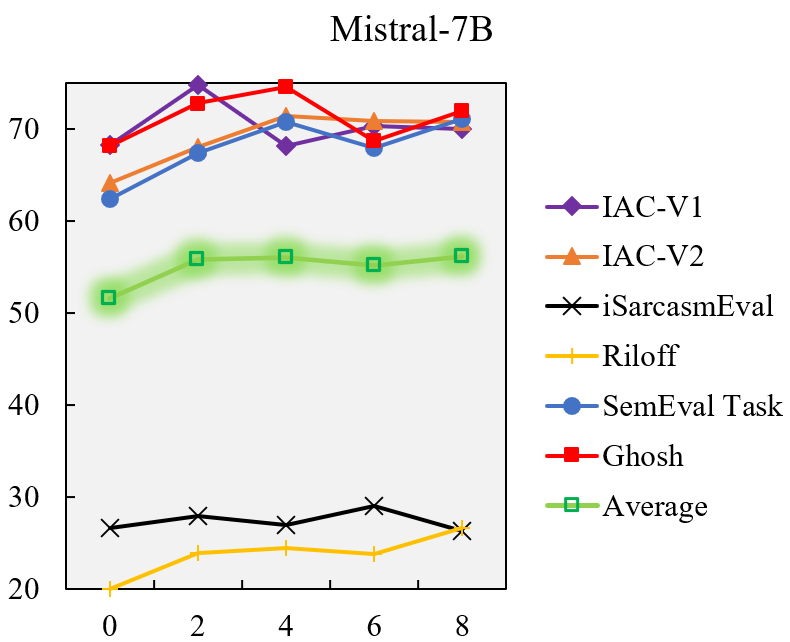}}
    \\
    \subfloat[]{\includegraphics[width=2.6in]{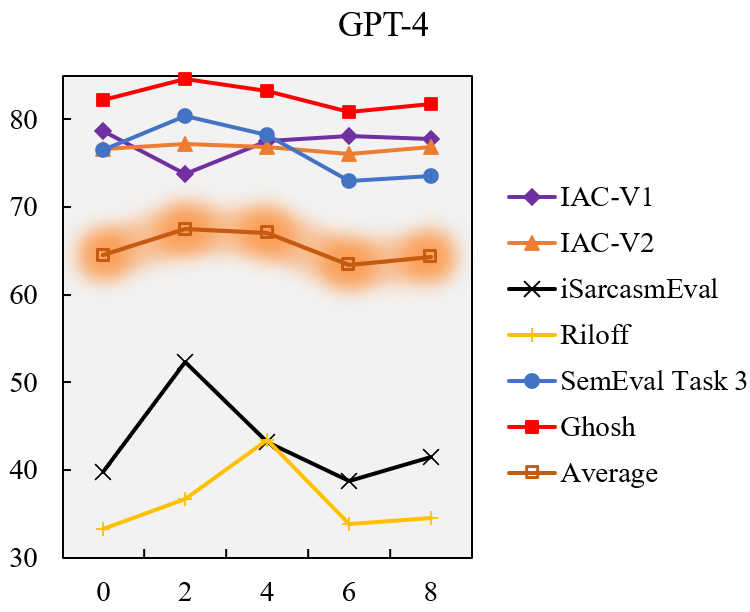}}
\caption{The F1 results of Mistral and GPT-4 using different numbers of demonstrations.} \label{fig:Fewshot}  
\end{figure}
\subsection{Few-Shot vs. Zero-Shot}
\begin{figure*}[t]
\centering
\includegraphics[width=6.3in]{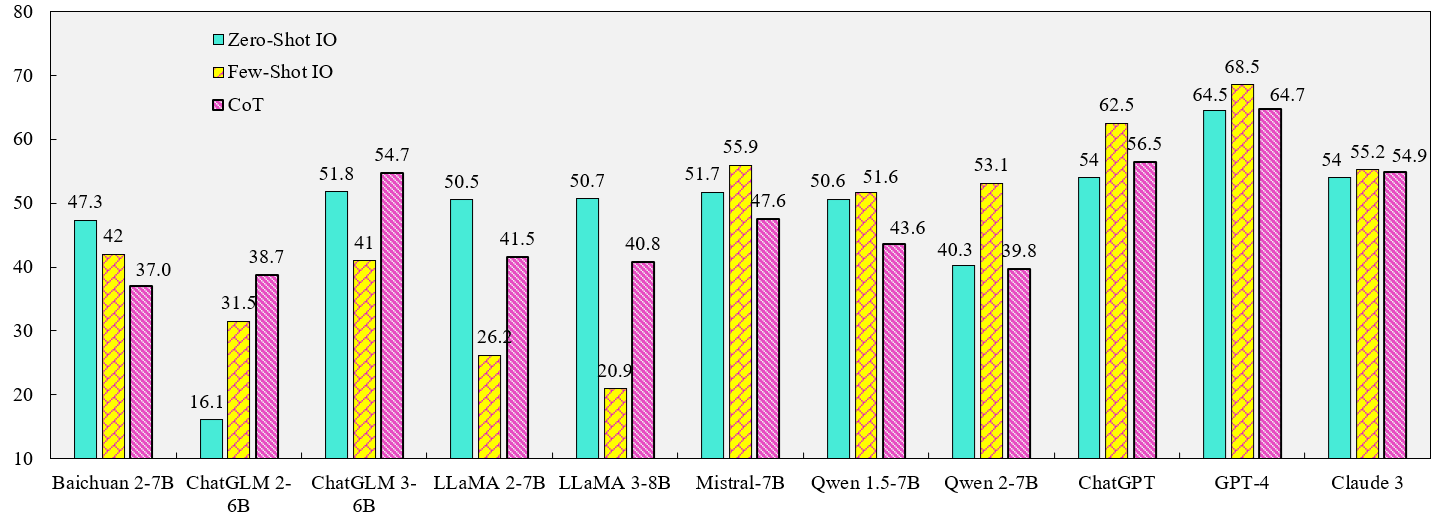}
\caption{The comparison between zero-shot IO prompting, few-shot IO prompting and CoT prompting.} \label{fig:k-shot}
\end{figure*}

Since the above experiments are mainly based on a zero-shot IO setting, we are curious of whether the conclusions also apply in a few-shot scenario. Therefore, we perform few-shot experiments to evaluate whether the LLMs can perform better with a limited number of examples. We show the main results in Table~\ref{tab:fewshotbaseline} and Fig.~\ref{fig:k-shot}, we sample $k=2$ examples (including one sarcastic example and one non-sarcastic example) from the training set via KNN search.
\begin{table*}[t!]
\centering
\caption{Performance on six datasets. For LLMs, all strategies are based on a \textbf{few-shot IO setting}. \textbf{\textcolor{blue}{Blue}} indicates the best results across LLMs.}
\label{tab:fewshotbaseline}
\scalebox{0.82}{
\begin{tabular}{cl|cccc|cccc|cccc|c}
\midrule[1pt]
                                                   &                         & \multicolumn{4}{c}{\textbf{IAC-V1}}                                     & \multicolumn{4}{c}{\textbf{IAC-V2}}                                     & \multicolumn{4}{c}{\textbf{iSarcasmEval}}                               &                                       \\ \cline{3-14}
                                                   
\multirow{-2}{*}{\textbf{Paradigm}}  & \multirow{-2}{*}{\textbf{Model}} & \textbf{Acc}          & \textbf{P}          & \textbf{R}           & \textbf{F1}            & \textbf{Acc}          & \textbf{P}          & \textbf{R}           & \textbf{F1}   & \textbf{Acc}          & \textbf{P}          & \textbf{R}           & \textbf{F1}           & \multirow{-2}{*}{\textbf{Avg. of F1}}           \\\midrule[1pt]

                            & RoBERTa                 & \textbf{\underline{70.1}}          & \textbf{\underline{70.0}}          & \textbf{\underline{70.1}}           & \textbf{\underline{69.9}}          & 80.7          & 80.9          & 80.8           & 80.7          & 78.9          & 66.4          & 57.5           & 63.5          & \cellcolor[HTML]{E5FBBB}71.3          \\

\multirow{-2}{*}{PLMs}             & DC-Net-RoBERTa          & 69.3          & 69.7          & 69.3           & 69.1          & \textbf{\underline{81.7}}          & \textbf{\underline{81.7}}          & \textbf{\underline{81.7}}           & \textbf{\underline{81.7}}          & \textbf{\underline{79.5}}             & \textbf{\underline{67.1}}             & \textbf{\underline{58.3}}              & \textbf{\underline{64.0}}             & \cellcolor[HTML]{E5FBBB}\textbf{\underline{71.5}}          \\
 \hline
 
                          &               Baichuan 2-7B            & 46.5                 & 57.5                 & 38.7                 & 46.3                 & 53.5                 & 53.9                 & 51.6                 & 52.7                 & 61.2                 & 13.9                 & 39.5                 & 20.5                 & \cellcolor[HTML]{E5FBBB}39.8          \\
                          
& ChatGLM 2-6B            & 46.3                 & 61.3                 & 26.2                 & 36.7                 & 52.3                 & 54.2                 & 31.9                 & 40.1                 & 58.2                 & 14.6                 & 47.4                 & 22.4                 & \cellcolor[HTML]{E5FBBB}33.1          \\

& ChatGLM 3-6B             & 56.1                 & 73.4                 & 41.1                 & 52.7                 & 66.2                 & 69.6                 & 58.1                 & 63.3                 & 73.6                 & 21.1                 & 39.5                 & 27.5                 & \cellcolor[HTML]{E5FBBB}47.8          \\

& LLaMA 2-7B               & 46.0                 & 55.8                 & 44.8                 & 50.0                 & 45.9                 & 45.8                 & 42.1                 & 43.9                 & 63.9                 & 10.2                 & 23.7                 & 14.3                 & \cellcolor[HTML]{E5FBBB}36.1          \\

& LLaMA 3-8B               & 43.4                 & 61.5                 & 12.9                 & 21.3                 & 56.4                 & 68.1                 & 24.7                 & 36.3                 & 84.6                 & 21.4                 & 7.9                 & 11.5                 & \cellcolor[HTML]{E5FBBB}23.0          \\

& Mistral-7B              & 65.2                 & 65.7                 & 86.7                 & 74.8       & 57.4                 & 54.6                 & 90.7                 & 68.1                 & 51.8                 & 17.3                 & 73.7                 & 28.0                 & \cellcolor[HTML]{E5FBBB}57.0          \\

& Qwen 1.5-7B            & 60.9                 & 62.5                 & 85.5                 & 72.2                 & 52.1                 & 51.5                 & 80.6                 & 62.8                 & 31.4                 & 13.9                 & 84.2                 & 23.8                 & \cellcolor[HTML]{E5FBBB}52.9          \\

& Qwen 2-7B            & 61.5                 & 63.5                 & 84.7                 & 73.0 & 53.5                 & 53.3                 & 75.1                & 62.7                & 33.0  & 16.7                & 78.8                & 26.4                & \cellcolor[HTML]{E5FBBB}54.0          \\

& ChatGPT           & 69.4                 & 74.3                 & 72.1                 & 73.2                 & 72.2                & 67.8                & 83.1                & 75.1                 & 76.1  & 34.7                 & 85.3                & 49.2                & \cellcolor[HTML]{E5FBBB}65.9 \\

& GPT-4 Turbo         & \textbf{\textcolor{blue}{\underline{73.3}}}        & \textbf{\textcolor{blue}{\underline{75.4}}}        & 84.6                 & \textbf{\textcolor{blue}{\underline{79.6}}}                 & \textbf{\textcolor{blue}{\underline{74.5}}}        & \textbf{\textcolor{blue}{\underline{70.0}}}        & 86.2                 & \textbf{\textcolor{blue}{\underline{77.2}}}        & \textbf{\textcolor{blue}{\underline{79.3}}}        & \textbf{\textcolor{blue}{\underline{37.0}}}        & 89.5                 & \textbf{\textcolor{blue}{\underline{52.3}}}        & \cellcolor[HTML]{E5FBBB}\textbf{\textcolor{blue}{\underline{69.7}}} \\
                                                       
\multirow{-11}{*}{LLMs$_{+\text{\textbf{k-shot~IO}}}$}    & Claude-3-haiku          & 63.5                 & 63.9                 & \textbf{\textcolor{blue}{\underline{89.1}}}        & 74.4                 & 61.0                 & 56.5                 & \textbf{\textcolor{blue}{\underline{96.6}}}        & 71.3                 & 36.5                 & 16.4                 & \textbf{\textcolor{blue}{\underline{97.4}}}        & 28.0                 & \cellcolor[HTML]{E5FBBB}57.9          \\ 
\midrule[1pt]

                                                     &                         & \multicolumn{4}{c}{\textbf{Riloff}}                                     & \multicolumn{4}{c}{\textbf{SemEval Task 3}}                            & \multicolumn{4}{c}{\textbf{Ghosh}} &                                       \\ \cline{3-14}
                                                     
\multirow{-2}{*}{\textbf{Paradigm}}  & \multirow{-2}{*}{\textbf{Model}} & \textbf{Acc}          & \textbf{P}          & \textbf{R}           & \textbf{F1}            & \textbf{Acc}          & \textbf{P}          & \textbf{R}           & \textbf{F1}   & \textbf{Acc}          & \textbf{P}          & \textbf{R}           & \textbf{F1}           & \multirow{-2}{*}{\textbf{Avg. of F1}}           \\\midrule[1pt]

                                                    & RoBERTa                 & \textbf{\underline{84.0  }}    & 64.0      & \textbf{\underline{75.0}}      & \textbf{\underline{69.0}}     & \textbf{\underline{75.0}}      & 64.0      & \textbf{\underline{84.0}}       & \textbf{\underline{72.0}}      & 80.3             & 80.3             & 79.8              & 80.0            & \cellcolor[HTML]{E5FBBB}\textbf{\underline{73.7}}             \\

\multirow{-2}{*}{PLMs}            & DC-Net-RoBERTa          & 82.6      & 67.4      & 70.3      & 67.8            & 70.9             & \textbf{\underline{69.7}}            & 68.3              & 68.7             & 82.2           &81.0             & 81.7              & 81.3             & \cellcolor[HTML]{E5FBBB}72.6             \\ \hline

                                                   & Baichuan 2-7B            & 52.2                 & 9.3                  & 50.0                 & 15.6                 & 53.7                 & 44.7                 & 71.1                 & 54.9                 & 58.5                 & 57.2                 & 67.3                 & 61.9                 & \cellcolor[HTML]{E5FBBB}44.1          \\

& ChatGLM 2-6B            & 51.3                 & 4.1                  & 20.0                 & 6.8                  & 52.6                 & 40.3                 & 40.8                 & 40.6                 & 45.1                 & 44.4                 & 39.6                 & 41.9                 & \cellcolor[HTML]{E5FBBB}29.8          \\

& ChatGLM 3-6B             & \textbf{\textcolor{blue}{\underline{81.4}}}                 & 7.7                  & 10.0                 & 8.7                  & 69.6                 & 63.3                 & 55.9                 & 59.4                 & 53.7                 & 59.1                 & 24.0                 & 34.1                 & \cellcolor[HTML]{E5FBBB}34.1          \\

& LLaMA 2-7B               & 66.4                 & 3.3                  & 10.0                 & 5.0                  & 46.9                 & 30.0                 & 25.4                 & 27.5                 & 45.3                 & 34.8                 & 10.8                 & 16.5                 & \cellcolor[HTML]{E5FBBB}16.3          \\

& LLaMA 3-8B               & 9.7                 & 8.9                  & \textbf{\textcolor{blue}{\underline{100.0}}}                 & 16.4                  & 60.8                 & 51.7                 & 20.0                 & 28.8                 & 51.3                 & 63.3                 & 6.2                 & 11.3                 & \cellcolor[HTML]{E5FBBB}18.8          \\

&  Mistral-7B              & 66.4                 & 15.0                 & 60.0                 & 24.0                 & 63.9                 & 52.5                 & 94.2                 & 67.4                 & 71.6                 & 69.9                 & 76.0                 & 72.8                 & \cellcolor[HTML]{E5FBBB}54.7          \\

& Qwen 1.5-7B            & 27.4                 & 10.9                 & \textbf{\textcolor{blue}{\underline{100.0}}}       & 19.6                 & 53.4                 & 45.9                 & 98.1                 & 62.6                 & 55.7                 & 53.2                 & 95.9                 & 68.4                 & \cellcolor[HTML]{E5FBBB}50.2          \\

& Qwen 2-7B            & 40.2                 & 16.8                 & 89.7                 & 20.8                 & 54.7                 & 50.3                 & 95.8                & 65.7                 & 57.9  & 55.6                 & 93.2                & 69.8                & \cellcolor[HTML]{E5FBBB}52.1          \\

& ChatGPT           & 67.5                & 20.0                 & 84.9               & 30.5                 & 68.9                 & 60.9                & 92.6                & 71.2                 & 76.8  & 72.3                 & 86.2              & 75.4                & \cellcolor[HTML]{E5FBBB}59.0          \\

& GPT-4 Turbo         & 72.6        & \textbf{\textcolor{blue}{\underline{23.1}}}        & 90.0                 & \textbf{\textcolor{blue}{\underline{36.7}}}        & \textbf{\textcolor{blue}{\underline{81.1}}}        & \textbf{\textcolor{blue}{\underline{68.3}}}        & 97.7                 & \textbf{\textcolor{blue}{\underline{80.4}}}        & \textbf{\textcolor{blue}{\underline{83.9}}}        & \textbf{\textcolor{blue}{\underline{80.7}}}        & 88.9                 & \textbf{\textcolor{blue}{\underline{84.6}}}        & \cellcolor[HTML]{E5FBBB}\textbf{\textcolor{blue}{\underline{67.2}}} \\

\multirow{-11}{*}{LLMs$_{+\text{\textbf{k-shot~IO}}}$}     & Claude-3-haiku          & 36.3                 & 12.2                 & \textbf{\textcolor{blue}{\underline{100.0}}}       & 21.7                 & 49.9                 & 44.2                 & \textbf{\textcolor{blue}{\underline{99.7}}}        & 61.2                 & 66.7                 & 60.2                 & \textbf{\textcolor{blue}{\underline{98.0}}}        & 74.6                 & \cellcolor[HTML]{E5FBBB}52.5 \\ \midrule[1pt]

\end{tabular}
}
\end{table*}

We can make two main observations. 
(1) In the few-shot IO prompting setting, GPT-4 consistently outperforms RoBERTa and DC-Net-RoBERTa on three datasets: IAC-V1, SemEval 2018 Task 3, and Ghosh. Specifically, GPT-4 achieves an F1 score of 79.6 on IAC-V1, which is significantly higher than RoBERTa's 69.9 and DC-Net-RoBERTa's 69.1. On SemEval 2018 Task 3, GPT-4 reaches an F1 score of 68.3, surpassing RoBERTa's 64.0 and DC-Net-RoBERTa's 67.8. Compared to the zero-shot setting, GPT-4 shows improvement on the remaining three datasets, indicating that with the benefit of a few examples, GPT-4 significantly enhances its performance. For example, on the iSarcasmEval dataset, GPT-4's F1 score increases from 39.8 in zero-shot to 52.3 in few-shot, marking a 31.4\% improvement. 

(2) A series of LLMs exhibit slight performance improvements compared to the zero-shot setting. For instance, ChatGPT's F1 score on the IAC-V1 dataset rises from 70.0 in zero-shot to 73.2 in few-shot, reflecting a 4.6\% improvement. Similarly, ChatGLM 3-6B shows an increase in its F1 score on IAC-V1 from 52.0 in zero-shot to 52.7 in few-shot, representing a 1.3\% gain. Mistral-7B displays a more notable improvement on the iSarcasmEval dataset, where its F1 score advances from 26.7 in zero-shot to 28.0 in few-shot, indicating a 2.9\% enhancement. Even though the extent of enhancement differs across models and datasets, the few-shot setup generally contributes to better outcomes. We can also see that 7B/6B LLMs has relatively based in-context learning ability and they cannot address too complex prompts. But for some business LLMs such as GPT-4, with larger model size, they have better in-context learning ability. 

\begin{figure}[ht!]
     \centering
    \subfloat[Zero-shot IO setting.]{\includegraphics[width=2.8in]{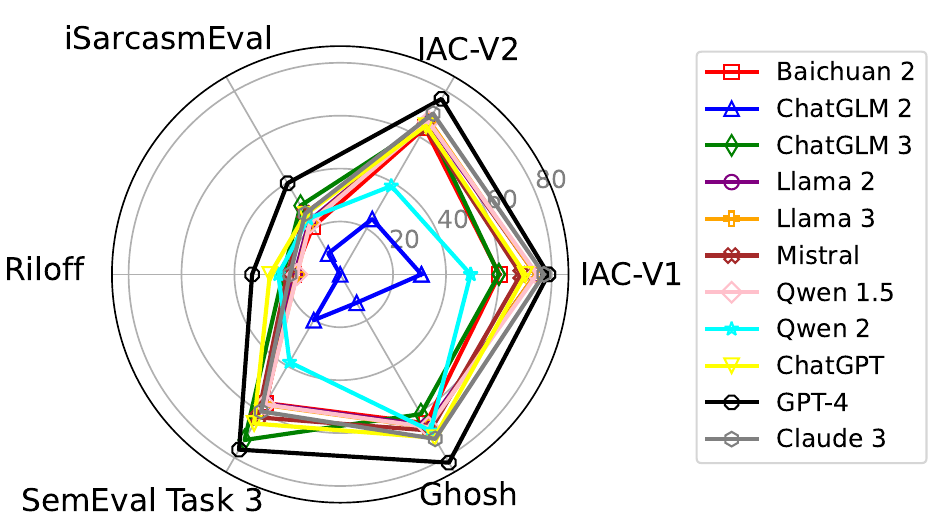}}
    \\
    \subfloat[Few-shot IO setting.]{\includegraphics[width=2.8in]{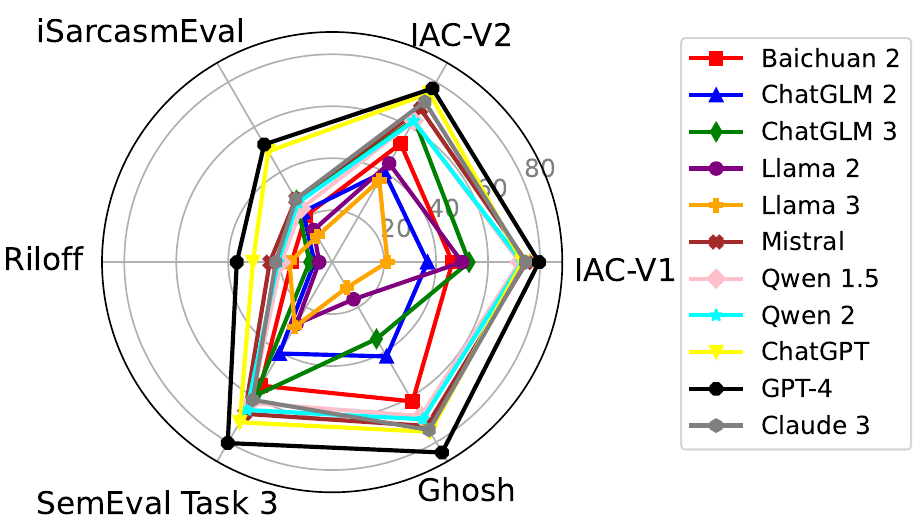}}
    \\
    \subfloat[CoT setting.]{\includegraphics[width=2.8in]{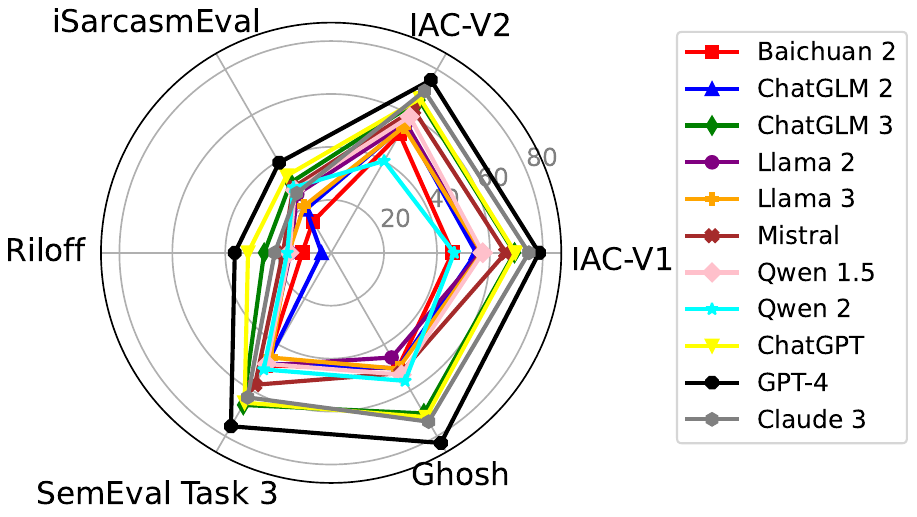}}
\caption{The radar illustration of all models' F1 performance across six datasets.}
\label{fig:rado}
\end{figure}

\subsection{Impact of Number of Demonstrations}\label{sec:impactof}
In few-shot settings, we investigate the impact of the number of demonstrations for the state-of-the-art LLM, namely whether GPT-4 could perform better or not if more contextual examples are provided.  We design five $k$-shot settings: zero-shot, two-shot, four-shot, six-shot, eight-shot. For each setting, we select  sample $k= \left \{ 0,2,4,6,8 \right \} $ examples that are similar to the test sample using the kNN sampling algorithm. 
The results are shown in Fig.~\ref{fig:Fewshot}.

As we can see, the number of demonstrations (k-shot) has a significant impact on the F1 scores for both Mistral and GPT-4 in the sarcasm classification task. For instance, Mistral shows an overall upward trend in F1 scores from 0-shot to 8-shot, increasing from 51.6 to 56.2, representing a total improvement of 8.7\% $\uparrow$. The most significant performance boost occurs at 2-shot, reaching 55.8, which is an 8.2\% increase compared to 0-shot. This shows that Mistral has greater sensitivity to the number of examples provided, benefiting more from additional examples.

In contrast, GPT-4 consistently outperforms Mistral across all shot numbers, with a baseline performance of 64.5 F1 score. GPT-4 achieves its best performance at 2-shot, with an F1 score of 67.5, showing a 4.6\% improvement over 0-shot. However, it experiences a slight decline in performance at 6-shot (63.4) and 8-shot (64.3). This indicates that GPT-4 may reach its peak performance with fewer examples and does not benefit as much from additional demonstrations.

These results suggest that both models reach their optimal performance with 2-4 samples, indicating that moderate few-shot learning may be more effective than using a large number of samples for sarcasm detection tasks. Sarcasm detection can vary significantly across textual contexts and datasets, potentially explaining the variations in optimal demonstration numbers.

\subsection{CoT Results}\label{sec:cotresults}
In few-shot settings, we investigate a widely used prompting approach, CoT. The experimental results are shown in Table~\ref{tab:cotbaseline} and Fig.~\ref{fig:k-shot}. Despite that CoT prompting has a significant advantage over standard IO prompting across logical, mathematical, and commonsense reasoning tasks, it may be not sufficient in solving sarcasm understanding tasks. For example, GPT-4 sees a decline in its average F1 score from 68.4 with IO prompting to 64.7 with CoT prompting, a 5.5\% decrease. Similarly, Mistral, which benefits from IO prompting with an F1 score of 55.8, drops to 47.5 with CoT prompting, reflecting a 14.88\% reduction. These examples highlight that while CoT can improve reasoning, it may introduce complexity that hinders models' ability to effectively detect and interpret sarcasm. Qwen 2 sees a 25.1\% reduction, from 53.1 to 39.7. These results indicate that CoT often complicates sarcasm understanding, negatively impacting performance. We present an intuitive comparison in Fig.~\ref{fig:rado}.

The reason is that sarcasm detection is often considered a holistic and non-rational cognitive process that does not conform to step-by-step logical reasoning. Sarcasm expression does not strictly conform to formal logical structures, such as the law of hypothetical syllogism (i.e., \textit{if} $\mathcal{A} \Rightarrow \mathcal{B} $ \textit{and} $\mathcal{B} \Rightarrow \mathcal{C} $, \textit{then} $\mathcal{A} \Rightarrow \mathcal{C} $). For example, ``\textit{Poor Alice has fallen for that stupid Bob; and that stupid Bob is head over heels for Claire; but don't assume for a second that Alice would like Claire}''. Another possible reason is that CoT does not explicitly harnesses the quintessential property of sarcasm (namely the contradiction between surface sentiment and true intention).
Hence, CoT prompting is less effective for sarcasm classification compared to the more straightforward IO approach. There is an urgent need to develop sarcasm understanding prompting approaches.

\begin{table*}[t!]
\centering
\caption{Performance on six datasets. For LLMs, all strategies are based on a \textbf{few-shot CoT setting}. \textbf{\textcolor{blue}{Blue}} indicates the best results across LLMs.}
\label{tab:cotbaseline}
\scalebox{0.82}{
\begin{tabular}{cl|cccc|cccc|cccc|c}
\midrule[1pt]
                                                   &                         & \multicolumn{4}{c}{\textbf{IAC-V1}}                                     & \multicolumn{4}{c}{\textbf{IAC-V2}}                                     & \multicolumn{4}{c}{\textbf{iSarcasmEval}}                               &                                       \\ \cline{3-14}
                                                   
\multirow{-2}{*}{\textbf{Paradigm}}  & \multirow{-2}{*}{\textbf{Model}} & \textbf{Acc}          & \textbf{P}          & \textbf{R}           & \textbf{F1}            & \textbf{Acc}          & \textbf{P}          & \textbf{R}           & \textbf{F1}   & \textbf{Acc}          & \textbf{P}          & \textbf{R}           & \textbf{F1}           & \multirow{-2}{*}{\textbf{Avg. of F1}}           \\\midrule[1pt]

                            & RoBERTa                 & \textbf{\underline{70.1}}          & \textbf{\underline{70.0}}          & \textbf{\underline{70.1}}           & \textbf{\underline{69.9}}          & 80.7          & 80.9          & 80.8           & 80.7          & 78.9          & 66.4          & 57.5           & 63.5          & \cellcolor[HTML]{E5FBBB}71.3          \\

\multirow{-2}{*}{PLMs}             & DC-Net-RoBERTa          & 69.3          & 69.7          & 69.3           & 69.1          & \textbf{\underline{81.7}}          & \textbf{\underline{81.7}}          & \textbf{\underline{81.7}}           & \textbf{\underline{81.7}}          & \textbf{\underline{79.5}}             & \textbf{\underline{67.1}}             & \textbf{\underline{58.3}}              & \textbf{\underline{64.0}}             & \cellcolor[HTML]{E5FBBB}\textbf{\underline{71.5}}          \\
 \hline
 
                          &              Baichuan 2-7B            & 46.8                 & 58.0                 & 37.9                 & 45.9                 & 52.5                 & 52.9                 & 50.7                 & 51.8                 & 57.9                 & 9.3                  & 26.3                 & 13.7                 & \cellcolor[HTML]{E5FBBB}37.1 \\

      &    ChatGLM 2-6B            & 51.3                 & 61.2                 & 50.0                 & 54.8                 & 55.0                 & 55.0                 & 56.7                 & 55.9                 & 57.2                 & 12.5                 & 39.5                 & 19.0                 & \cellcolor[HTML]{E5FBBB}43.2 \\

      &    ChatGLM 3-6B             & 60.7                 & 64.7                 & 74.6                 & 69.3                 & 59.5                 & 56.8                 & 80.5                 & 66.6                 & 49.8                 & 18.5                 & 86.8                 & 30.6                 & \cellcolor[HTML]{E5FBBB}55.5 \\

      &    LLaMA 2-7B               & 51.1                 & 59.9                 & 53.6                 & 56.6                 & 52.0                 & 51.8                 & 62.2                 & 56.6                 & 49.5                 & 15.8                 & 68.4                 & 25.6                 & \cellcolor[HTML]{E5FBBB}46.3 \\

      &    LLaMA 3-8B               & 52.5                 & 61.7                 & 53.2                 & 57.1                 & 52.3                 & 52.2                 & 57.8                 & 54.9                 & 55.5                 & 13.2                 & 44.7                 & 20.4                 & \cellcolor[HTML]{E5FBBB}44.1 \\

      &    Mistral-7B              & 59.0                 & 65.1                 & 66.9                 & 66.0                 & 57.1                 & 55.7                 & 71.0                 & 62.4                 & 59.9                 & 18.5                 & 63.2                 & 28.6                 & \cellcolor[HTML]{E5FBBB}52.3 \\

      &    Qwen 1.5-7B            & 52.5                 & 61.6                 & 53.6                 & 57.3                 & 55.7                    & 52.6                    & 64.3                    & 59.4   & 51.8                 & 16.9                 & 71.1                 & 27.3                 & \cellcolor[HTML]{E5FBBB}48.0    \\

      &    Qwen 2-7B            & 54.7                 & 42.5                &51.4                 & 46.5                    & 52.9                    &40.1                    & 56.4                    & 40.1                & 53.7 & 20.3 & 71.8   &28.5             & \cellcolor[HTML]{E5FBBB}38.3    \\

      &    ChatGPT            & 64.7                 &64.4                 & 81.5               & 69.6  & 61.9                   &56.4                    & 82.8                   & 67.3                    & 53.6                 & 15.6 & 87.7                & 33.6                 & \cellcolor[HTML]{E5FBBB}56.8    \\

      &    GPT-4-Turbo         & \textbf{\textcolor{blue}{\underline{72.2}}}                 & \textbf{\textcolor{blue}{\underline{72.4}}}                 & 85.9                 & \textbf{\textcolor{blue}{\underline{78.6}}}                 & \textbf{\textcolor{blue}{\underline{69.5}}}                 & \textbf{\textcolor{blue}{\underline{63.4}}}                 & 93.0                 & \textbf{\textcolor{blue}{\underline{75.4}}}                 & \textbf{\textcolor{blue}{\underline{65.9}}}                 & \textbf{\textcolor{blue}{\underline{25.4}}}                 & 86.8                 & \textbf{\textcolor{blue}{\underline{39.3}}}                 & \cellcolor[HTML]{E5FBBB}\textbf{\textcolor{blue}{\underline{64.4}}} \\

\multirow{-11}{*}{LLMs$_{+\text{\textbf{CoT}}}$}    & Claude-3-haiku          & 63.8                 & 63.9                 & \textbf{\textcolor{blue}{\underline{89.9}}}                 & 74.7                 & 59.8                 & 55.8                 & \textbf{\textcolor{blue}{\underline{95.8}}}                 & 70.6                 & 33.4                 & 15.2                 & \textbf{\textcolor{blue}{\underline{92.1}}}                 & 26.0                 & \cellcolor[HTML]{E5FBBB}57.1         \\ 
\midrule[1pt]

                                                     &                         & \multicolumn{4}{c}{\textbf{Riloff}}                                     & \multicolumn{4}{c}{\textbf{SemEval Task 3}}                            & \multicolumn{4}{c}{\textbf{Ghosh}} &                                       \\ \cline{3-14}
                                                     
\multirow{-2}{*}{\textbf{Paradigm}}  & \multirow{-2}{*}{\textbf{Model}} & \textbf{Acc}          & \textbf{P}          & \textbf{R}           & \textbf{F1}            & \textbf{Acc}          & \textbf{P}          & \textbf{R}           & \textbf{F1}   & \textbf{Acc}          & \textbf{P}          & \textbf{R}           & \textbf{F1}           & \multirow{-2}{*}{\textbf{Avg. of F1}}           \\\midrule[1pt]

                                                    & RoBERTa                 & \textbf{\underline{84.0  }}    & 64.0      & \textbf{\underline{75.0}}      & \textbf{\underline{69.0}}     & \textbf{\underline{75.0}}      & 64.0      & \textbf{\underline{84.0}}       & \textbf{\underline{72.0}}      & 80.3             & 80.3             & 79.8              & 80.0            & \cellcolor[HTML]{E5FBBB}\textbf{\underline{73.7}}             \\

\multirow{-2}{*}{PLMs}            & DC-Net-RoBERTa          & 82.6      & 67.4      & 70.3      & 67.8            & 70.9             & \textbf{\underline{69.7}}            & 68.3              & 68.7             & 82.2           &81.0             & 81.7              & 81.3             & \cellcolor[HTML]{E5FBBB}72.6             \\ \hline

                               &                             Baichuan 2-7B            & 56.6                 & 6.7                  & 30.0                 & 10.9                 & 53.7                 & 43.6                 & 56.6                 & 49.2                 & 50.7                 & 50.7                 & 50.1                 & 50.4                 & \cellcolor[HTML]{E5FBBB}36.7 \\

      &    ChatGLM 2-6B            & 52.2                 & 2.2                  & 10.0                 & 3.6                  & 54.3                 & 43.9                 & 54.0                 & 48.4                 & 50.1                 & 50.1                 & 50.8                 & 50.4                 & \cellcolor[HTML]{E5FBBB}34.1 \\

      &    ChatGLM 3-6B             & 47.8                 & 14.5                 & \textbf{\textcolor{blue}{\underline{100.0}}}                & 25.3                 & \textbf{\textcolor{blue}{\underline{88.1}}}                 & 53.3                 & 64.7                 & 66.4                 & \textbf{\textcolor{blue}{\underline{83.3}}}                 & 60.4                 & 64.4                 & 70.1                 & \cellcolor[HTML]{E5FBBB}53.9 \\

      &    LLaMA 2-7B               & 51.3                 & 9.1                  & 50.0                 & 15.4                 & 51.8                 & 42.2                 & 58.5                 & 49.1                 & 45.1                 & 45.2                 & 46.3                 & 45.7                 & \cellcolor[HTML]{E5FBBB}36.7 \\

      &    LLaMA 3-8B               & 54.0                 & 9.6                  & 50.0                 & 16.1                 & 51.1                 & 40.9                 & 51.8                 & 45.7                 & 49.9                 & 49.9                & 51.4                 & 50.6                 & \cellcolor[HTML]{E5FBBB}37.5 \\

      &    Mistral-7B              & 68.1                 & 11.8                 & 40.0                 & 18.2                 & 61.5                 & 51.1                 & 65.9                 & 57.6                 & 54.7                 & 55.1                 & 50.1                 & 52.5                 & \cellcolor[HTML]{E5FBBB}42.8 \\

      &    Qwen 1.5-7B            & 43.4                 & 9.1                  & 60.0                 & 15.8                 & 51.5                   & 43.1                    & 54.7                   & 48.5                    & 55.8                    & 55.7                    & 52.4                    &   53.2  & \cellcolor[HTML]{E5FBBB}39.2    \\
      
            &    Qwen 2-7B            & 49.9                 & 10.5                 & 66.0                 & 16.6                 & 56.7                 & 47.4                 & 60.2                 & 50.9                 & 61.4                 & 61.3                 & 57.6                 & 55.9                 & \cellcolor[HTML]{E5FBBB}41.1 \\
            
                  &    ChatGPT         & 59.6                 & 19.1                 & 86.4                & 31.4                 & 64.9                 & 53.4                 & 84.4                 & 65.4                 & 69.8                 & 64.3                 & 80.9                 & 71.7                   & \cellcolor[HTML]{E5FBBB}56.1   \\

      &    GPT-4-Turbo         & \textbf{\textcolor{blue}{\underline{69.0}}}                 & \textbf{\textcolor{blue}{\underline{22.2}}}                 & \textbf{\textcolor{blue}{\underline{100.0}}}                & \textbf{\textcolor{blue}{\underline{36.4}}}                 & 75.1                 & \textbf{\textcolor{blue}{\underline{61.8}}}                 & 97.7                 & \textbf{\textcolor{blue}{\underline{75.7}}}                 & 80.8                 & \textbf{\textcolor{blue}{\underline{74.5}}}                 & 93.6                 & \textbf{\textcolor{blue}{\underline{83.0}}}                 & \cellcolor[HTML]{E5FBBB}\textbf{\textcolor{blue}{\underline{65.0}}} \\

\multirow{-11}{*}{LLMs$_{+\text{\textbf{CoT}}}$}     & Claude-3-haiku          & 34.5                 & 11.9                 & \textbf{\textcolor{blue}{\underline{100.0}}}                & 21.3                 & 54.0                 & 46.2                 & \textbf{\textcolor{blue}{\underline{99.0                }}} & 63.1                 & 65.3                 & 59.3                 & \textbf{\textcolor{blue}{\underline{97.2}}}                 & 73.7                 & \cellcolor[HTML]{E5FBBB}52.7 \\ \midrule[1pt]

\end{tabular}
}
\end{table*}

\begin{figure*}[ht!]
    \centering
    \subfloat[IAC-V1]{\includegraphics[width=2.1in]{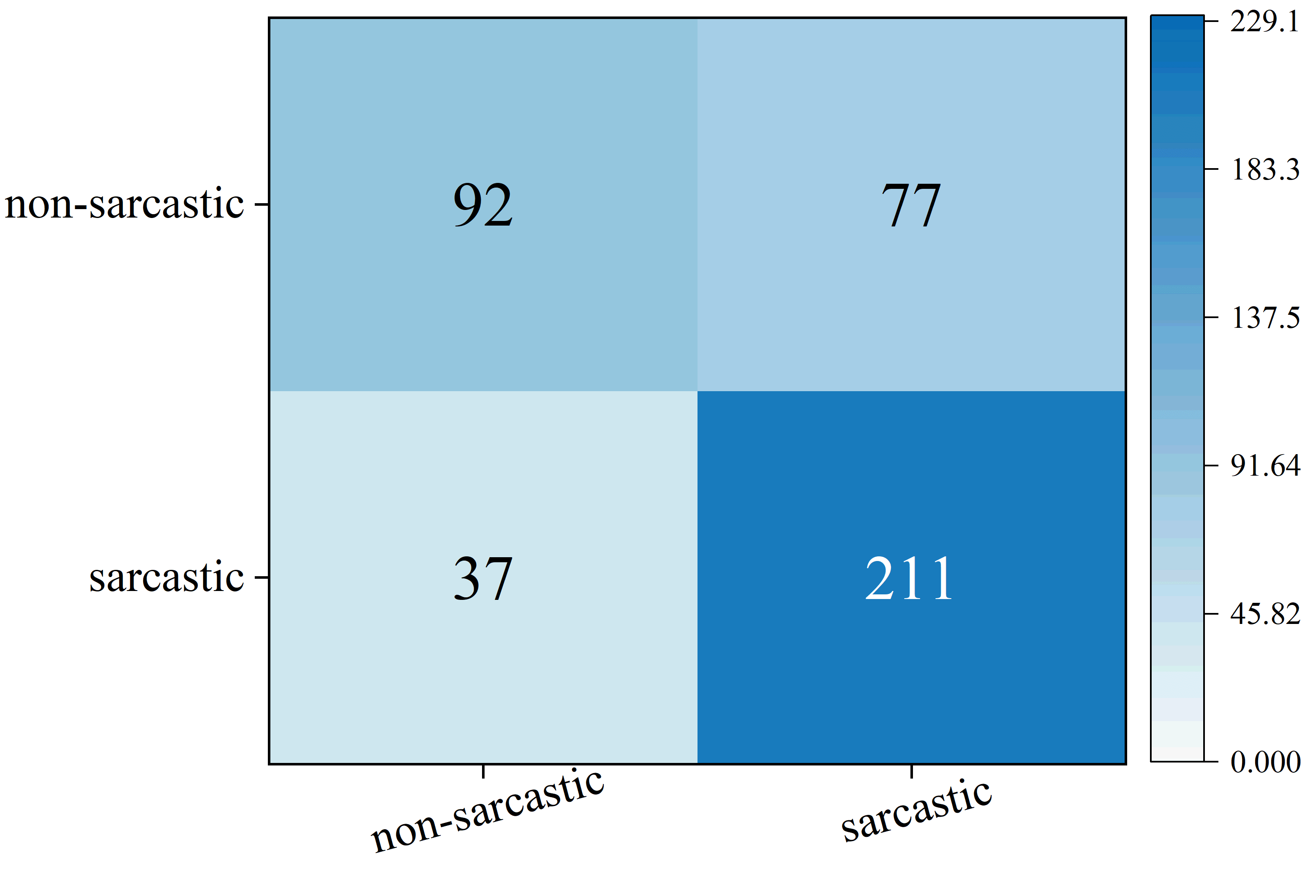}}
    \hfil 
    \subfloat[IAC-V2]{\includegraphics[width=2.1in]{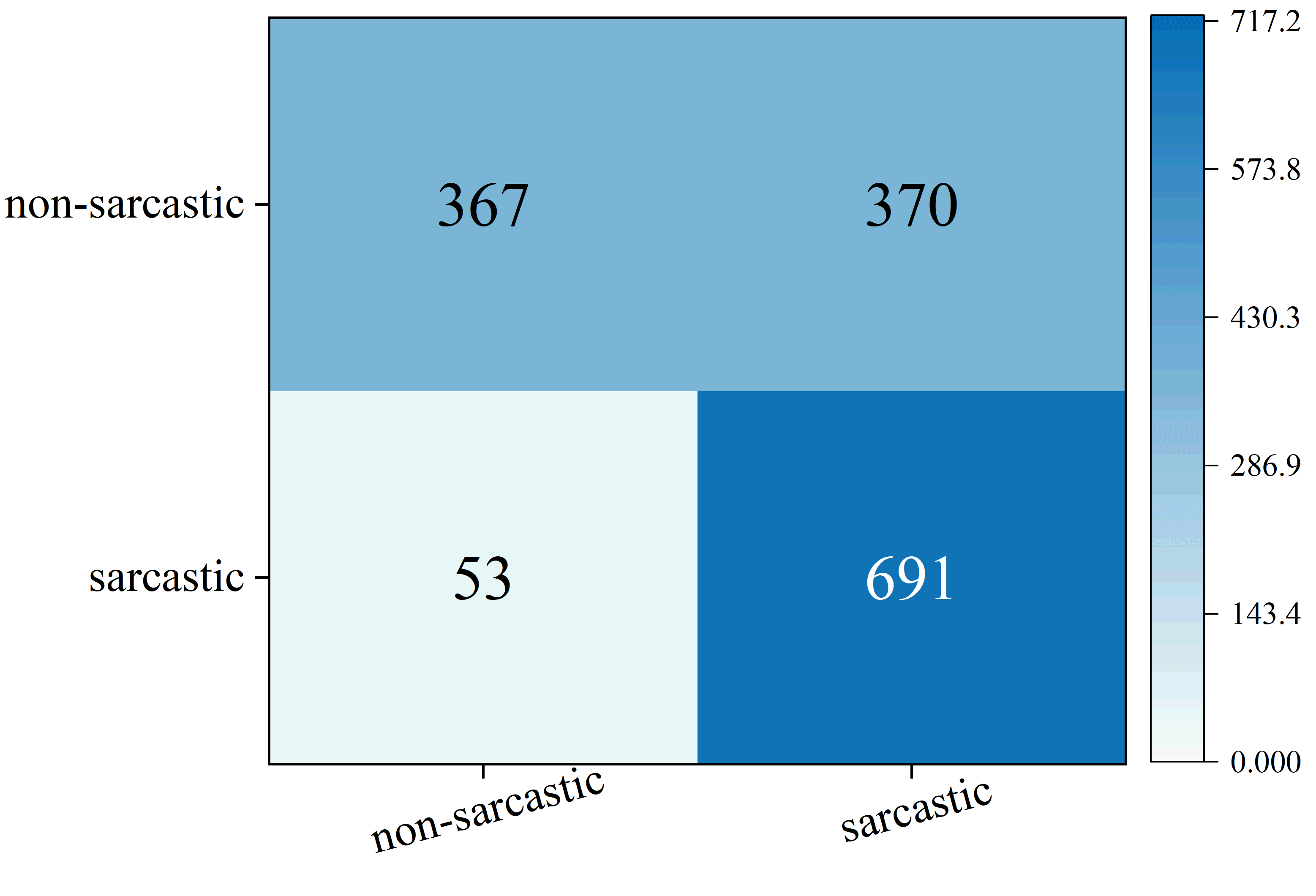}}%
    \hfil 
    \subfloat[Roliff]{\includegraphics[width=2.1in]{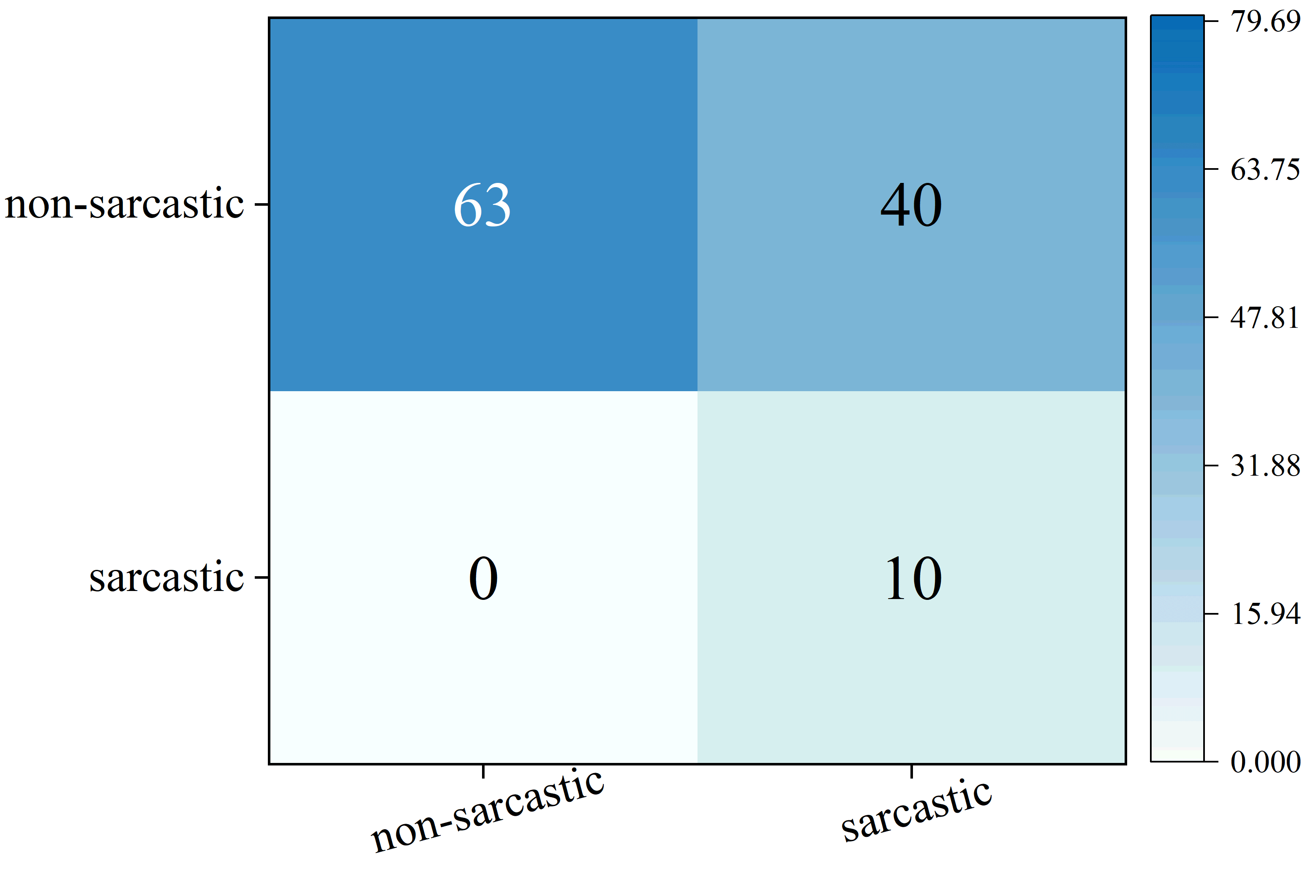}}
    \hfil 
    \subfloat[iSarcasmEval]{\includegraphics[width=2.1in]{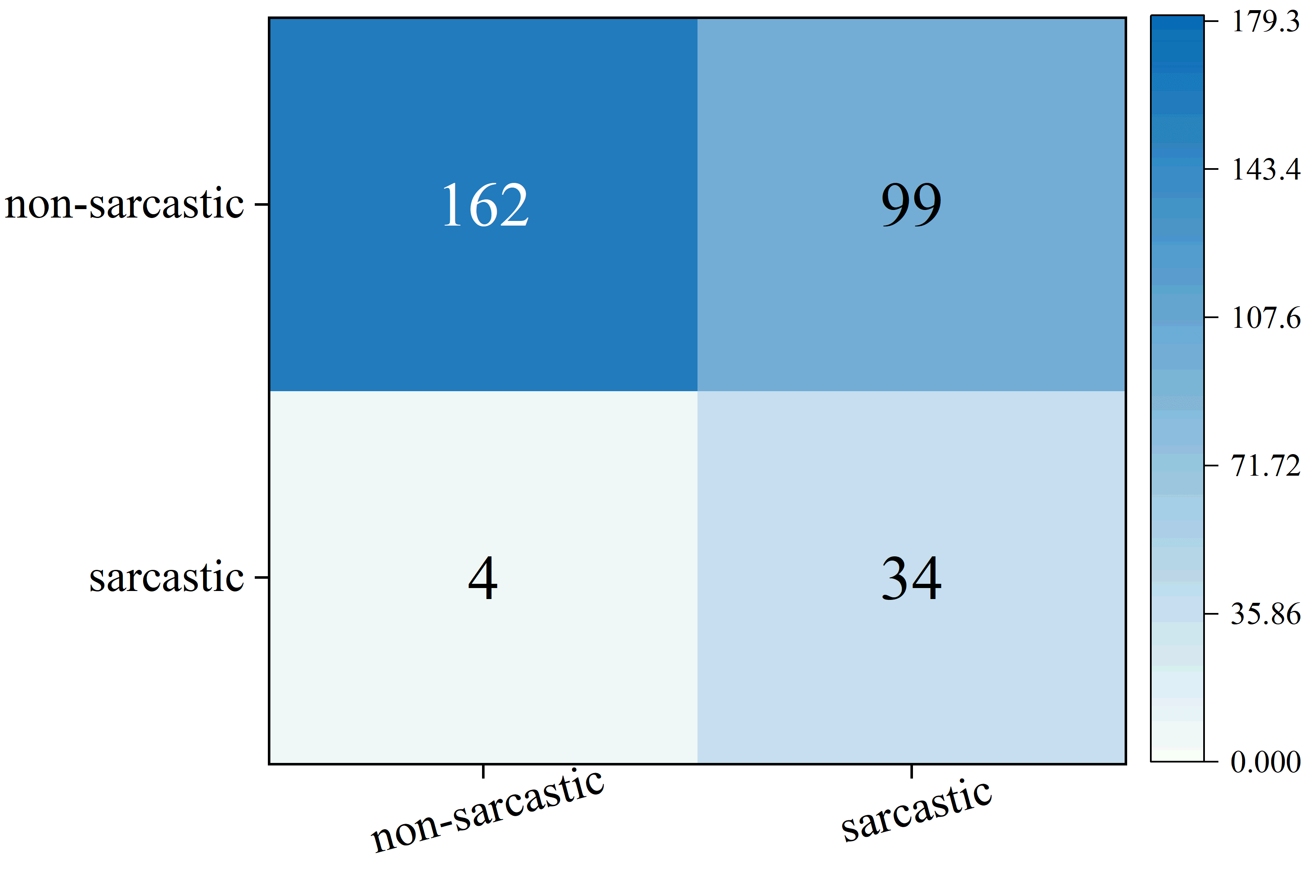}}
    \hfil 
    \subfloat[SemEval Task 3]{\includegraphics[width=2.1in]{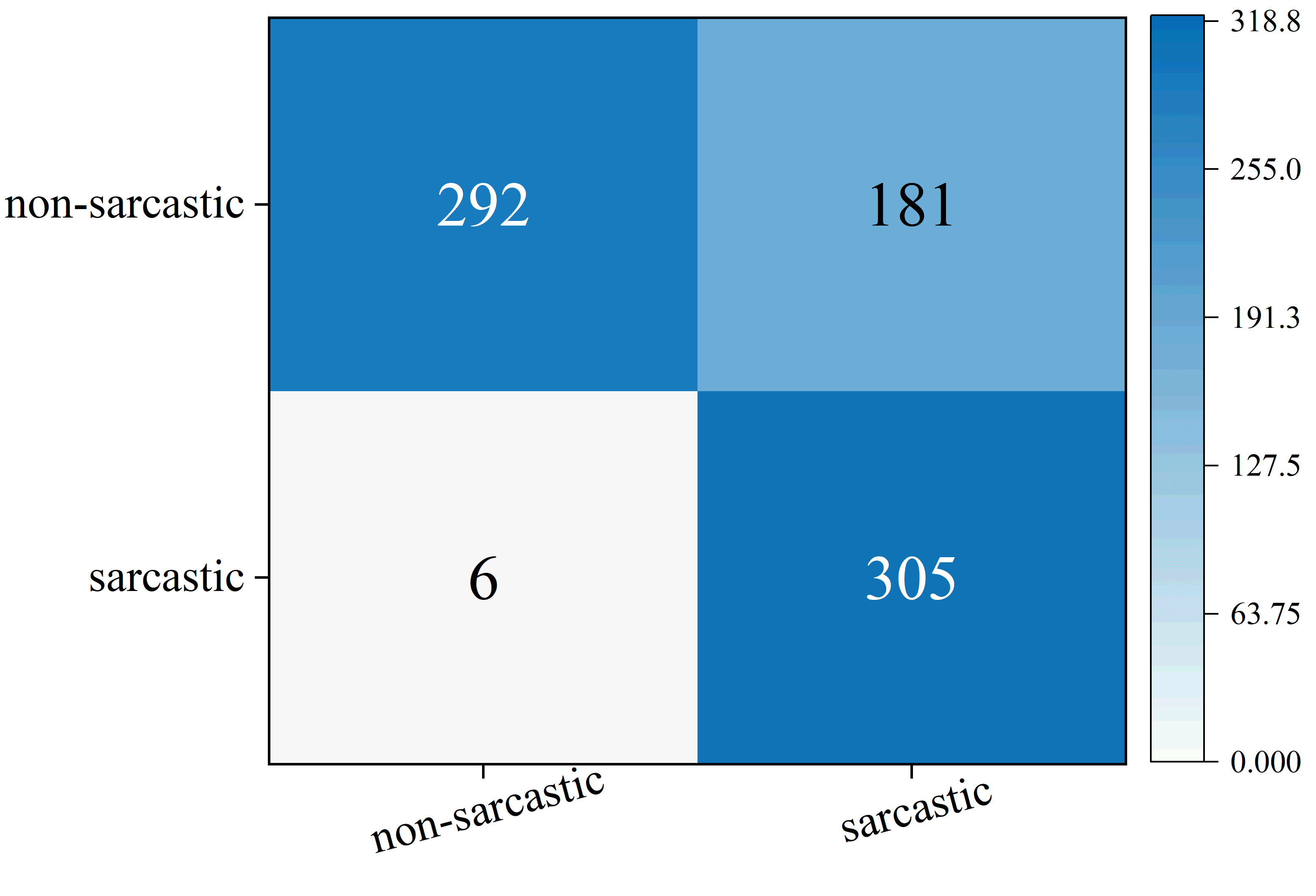}}
    \hfil
    \subfloat[Ghosh]{\includegraphics[width=2.1in]{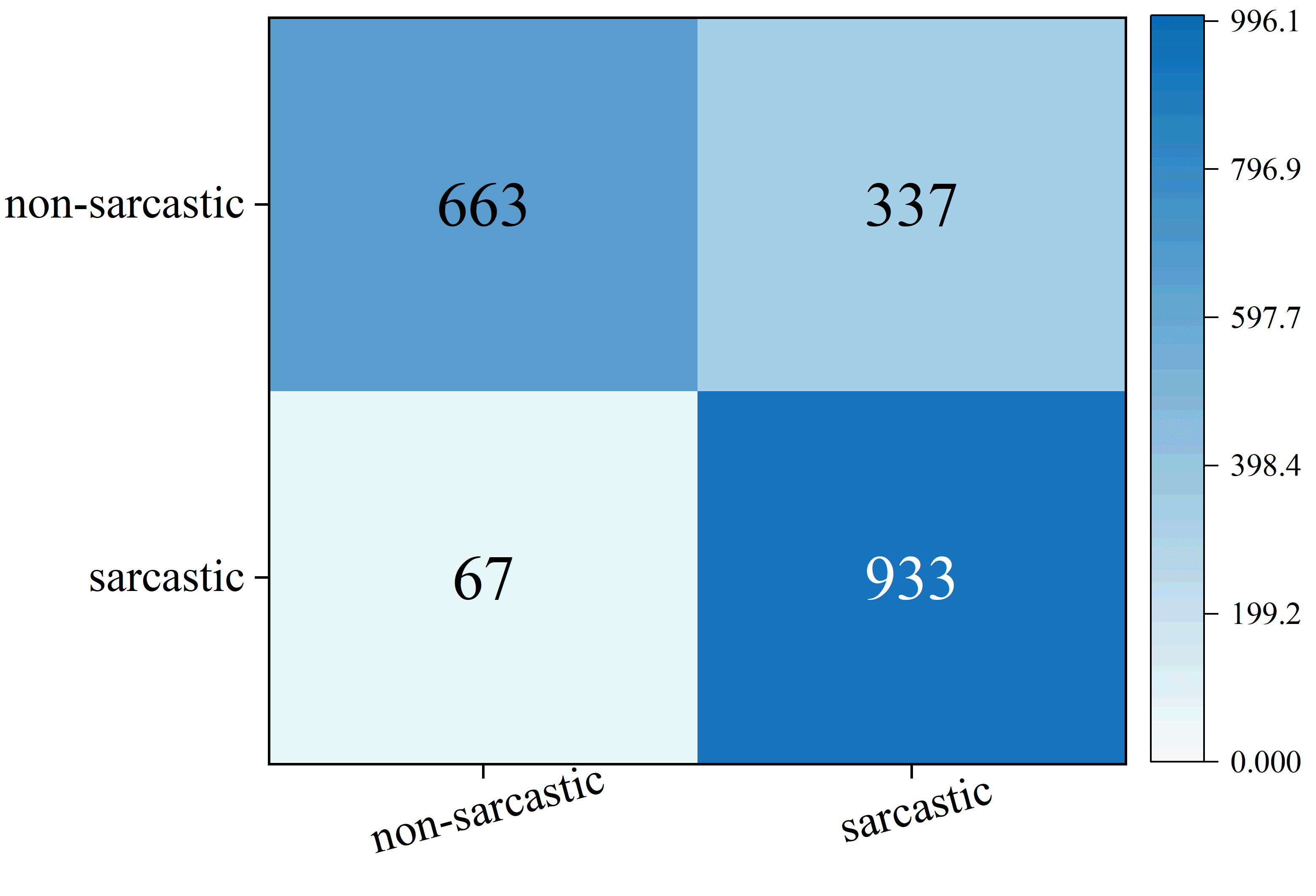}}
    \hfil 
    \caption{The normalized confusion matrices for GPT-4 across six datasets in the \textbf{zero-shot IO setting}.}\label{fig:confusionmatrix1}
\end{figure*}

\begin{figure*}[ht!]
    \centering
    \subfloat[IAC-V1]{\includegraphics[width=2.1in]{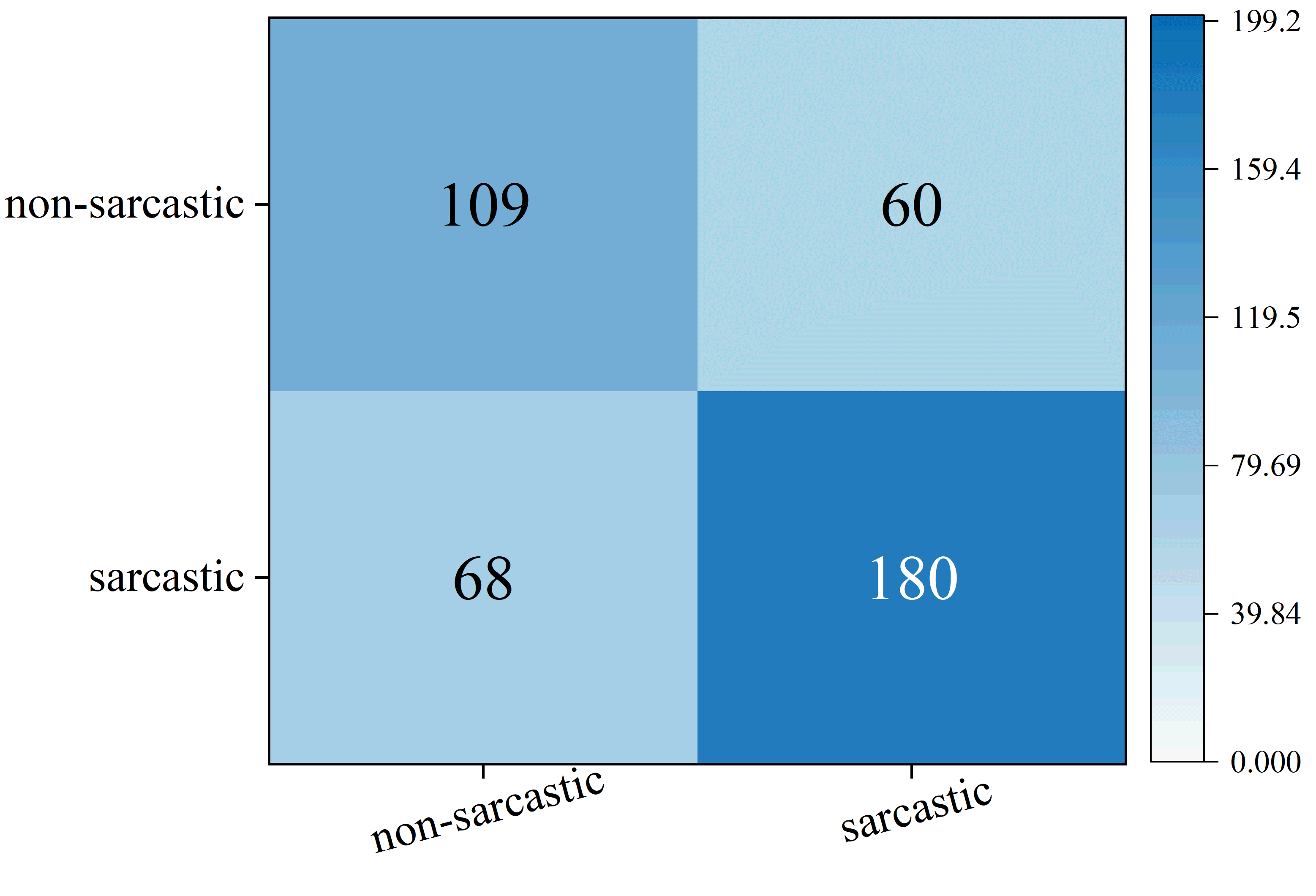}
    \label{fig: SinglePrompt}}
    \hfil 
    \subfloat[IAC-V2]{\includegraphics[width=2.1in]{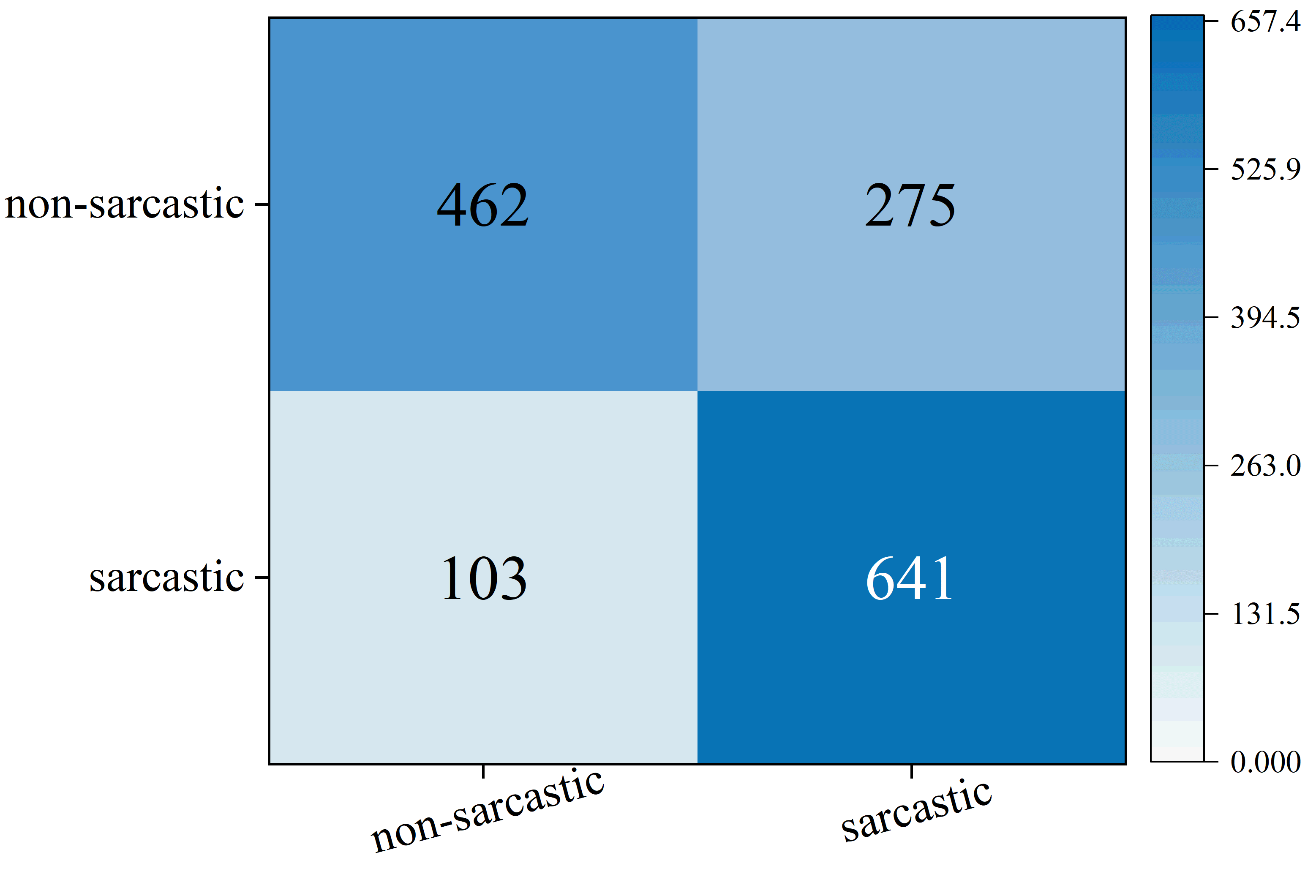}%
    \label{fig:MultiPrompt}}
    \hfil 
    \subfloat[Roliff]{\includegraphics[width=2.1in]{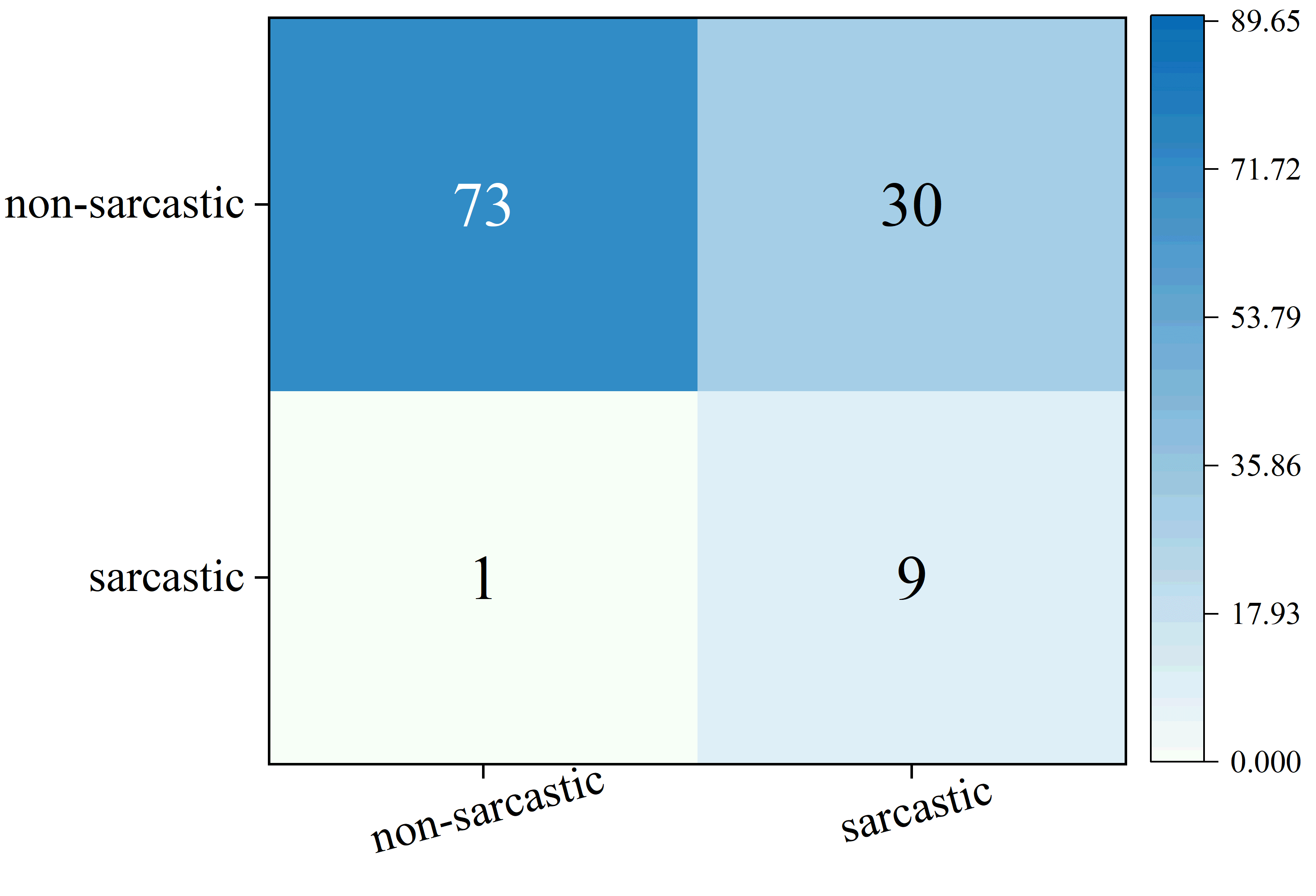}
    \label{fig: SinglePrompt}}
    \hfil 
    \subfloat[iSarcasmEval]{\includegraphics[width=2.1in]{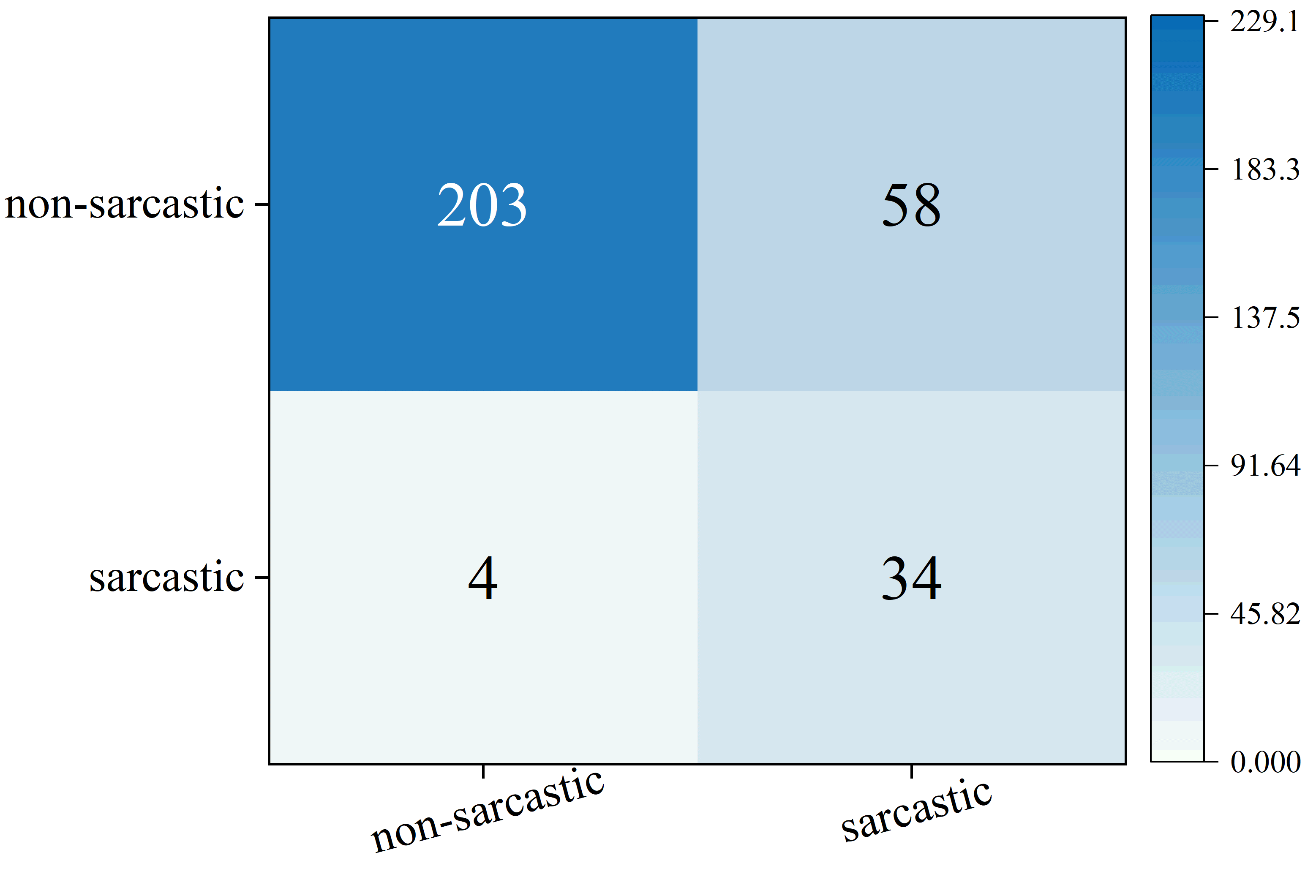}
    \label{fig: SinglePrompt}}
    \hfil 
    \subfloat[SemEval Task 3]{\includegraphics[width=2.1in]{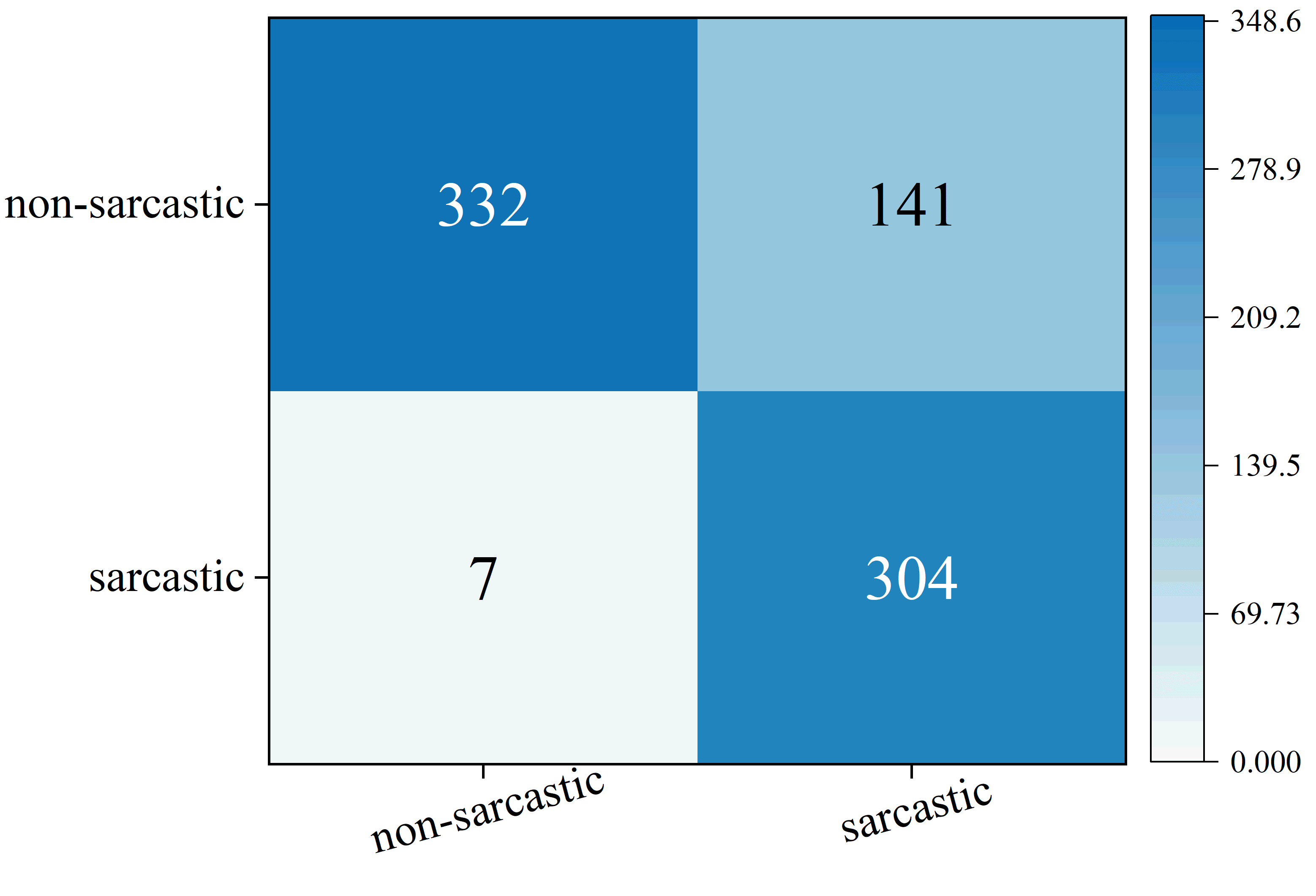}
    \label{fig: SinglePrompt}}
    \hfil
    \subfloat[Ghosh]{\includegraphics[width=2.1in]{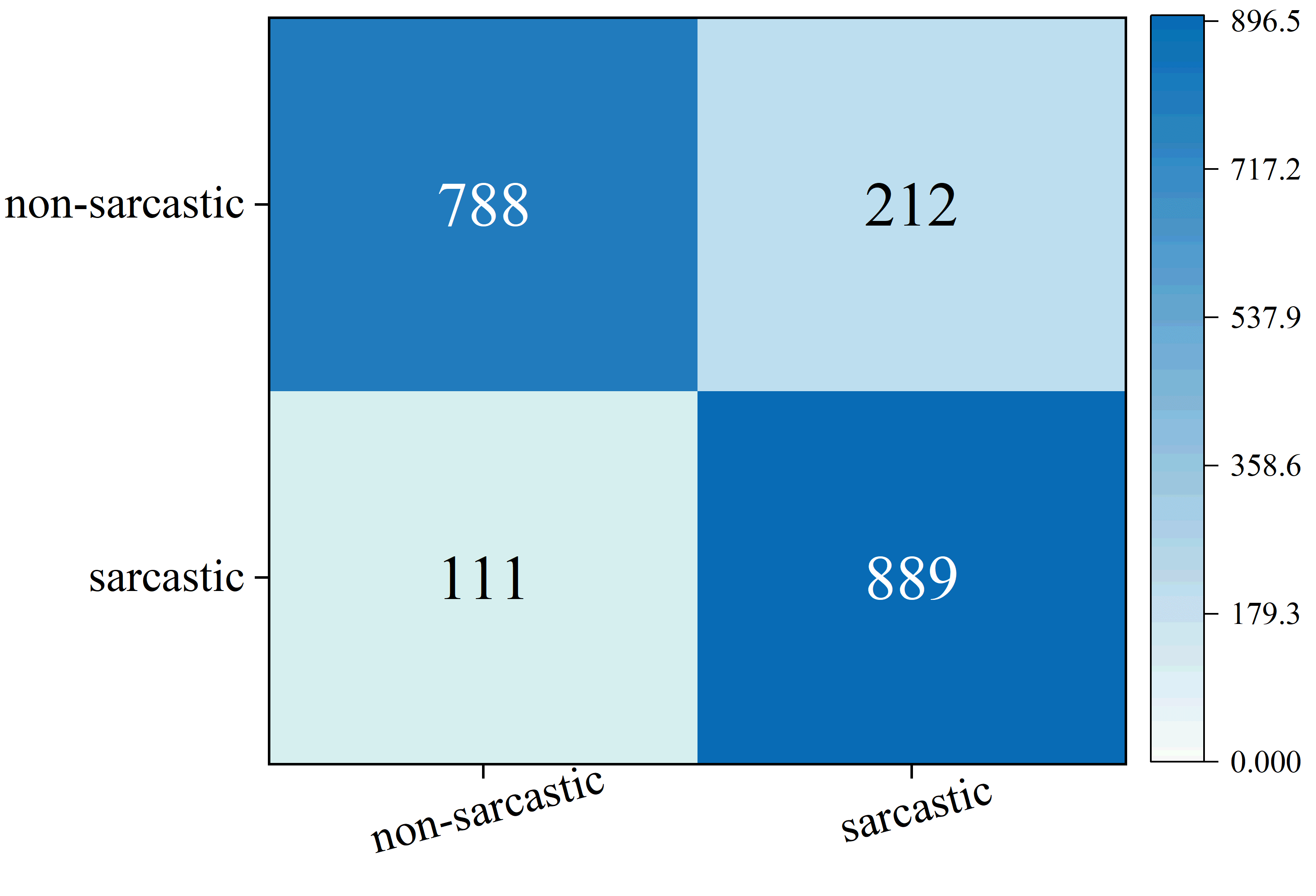}
    \label{fig: SinglePrompt}}
    \hfil 
\caption{The normalized confusion matrices for GPT-4 across six datasets in the \textbf{few-shot IO setting}.}
\label{fig:confusionmatrix2}
\end{figure*}
\begin{figure*}[!ht]
    \centering
    \subfloat[IAC-V1]{\includegraphics[width=2.1in]{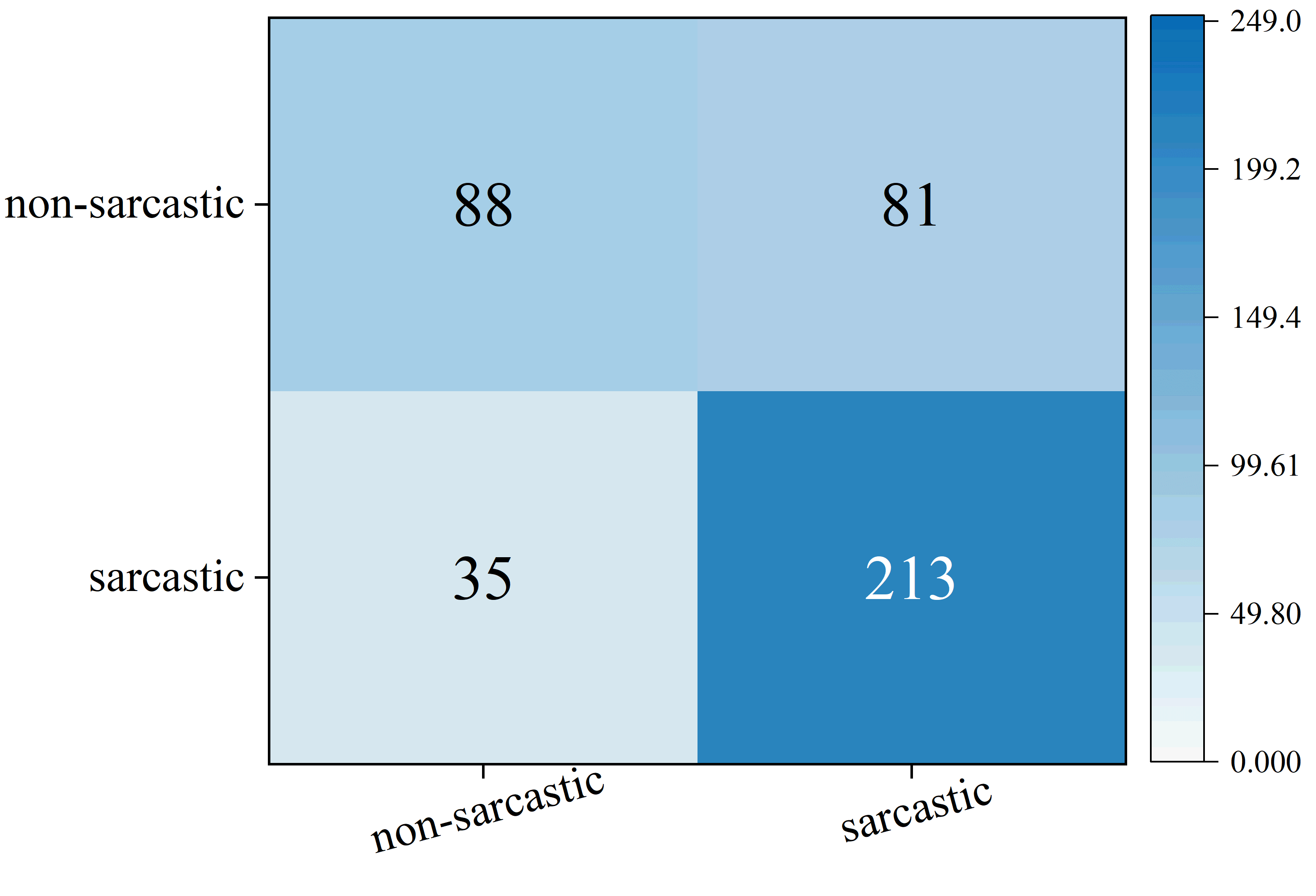}
    \label{fig: SinglePrompt}}
    \hfil 
    \subfloat[IAC-V2]{\includegraphics[width=2.1in]{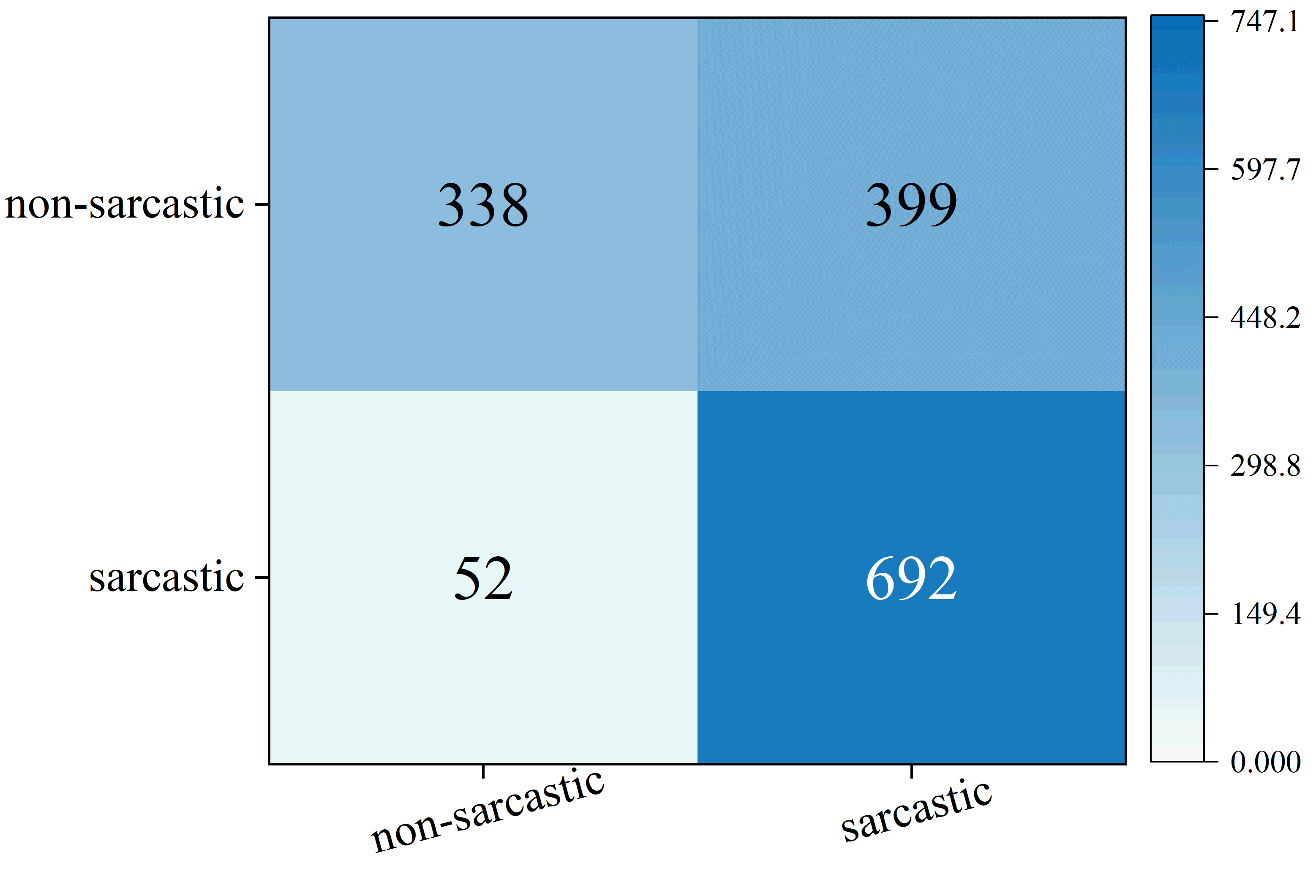}%
    \label{fig:MultiPrompt}}
    \hfil 
    \subfloat[Roliff]{\includegraphics[width=2.1in]{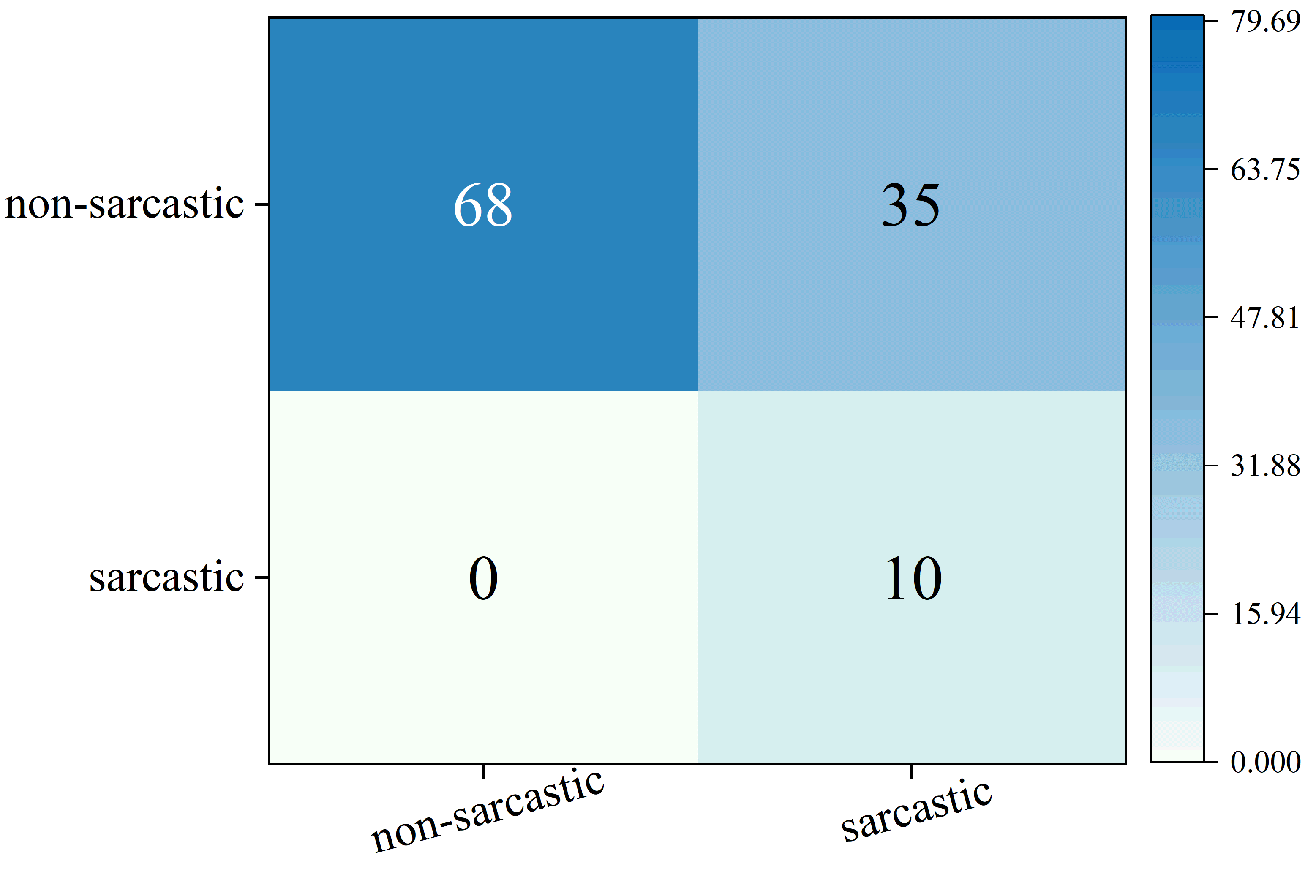}%
    \label{fig: SinglePrompt}}
    \hfil 
    \subfloat[iSarcasmEval]{\includegraphics[width=2.1in]{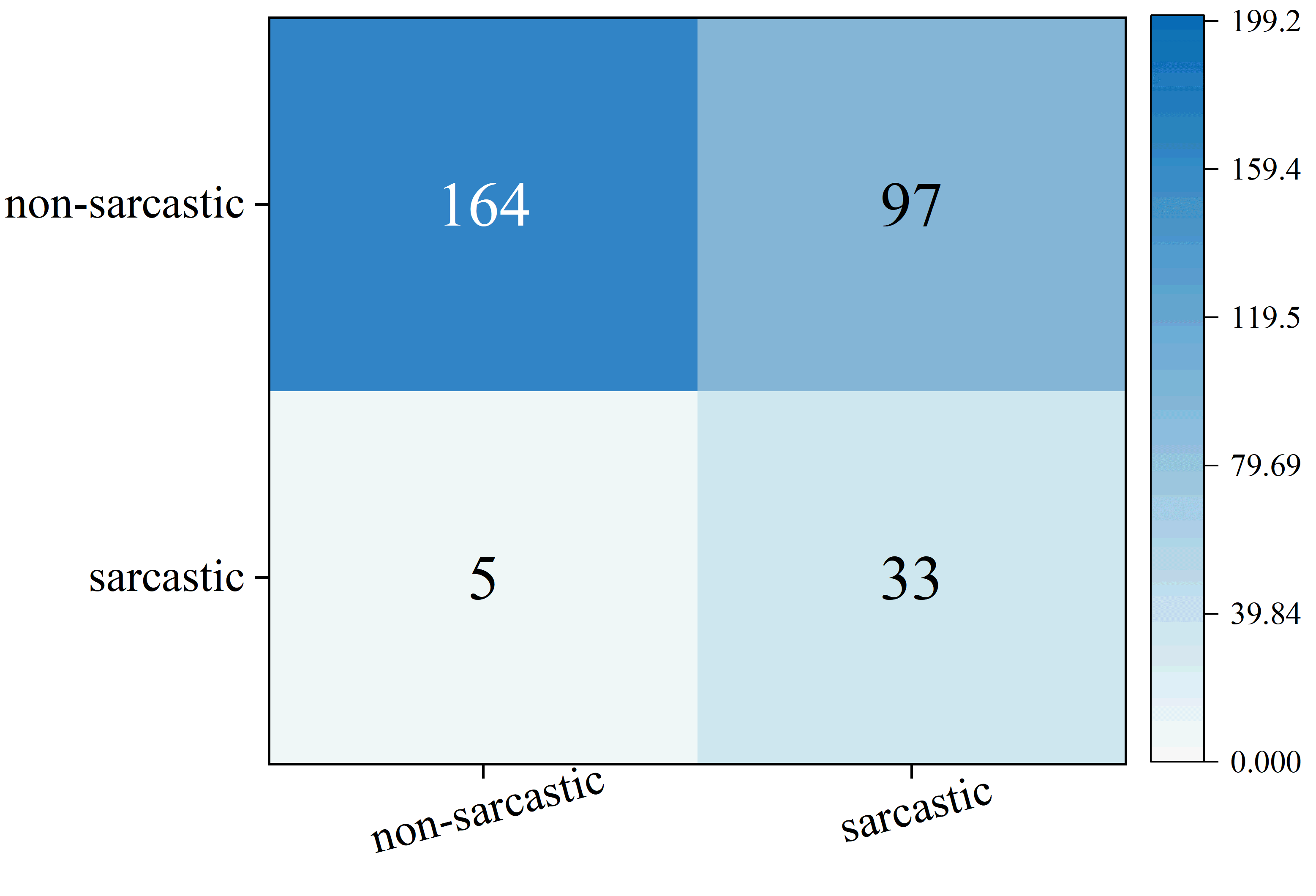}
    \label{fig: SinglePrompt}}
    \hfil 
    \subfloat[SemEval Task 3]{\includegraphics[width=2.1in]{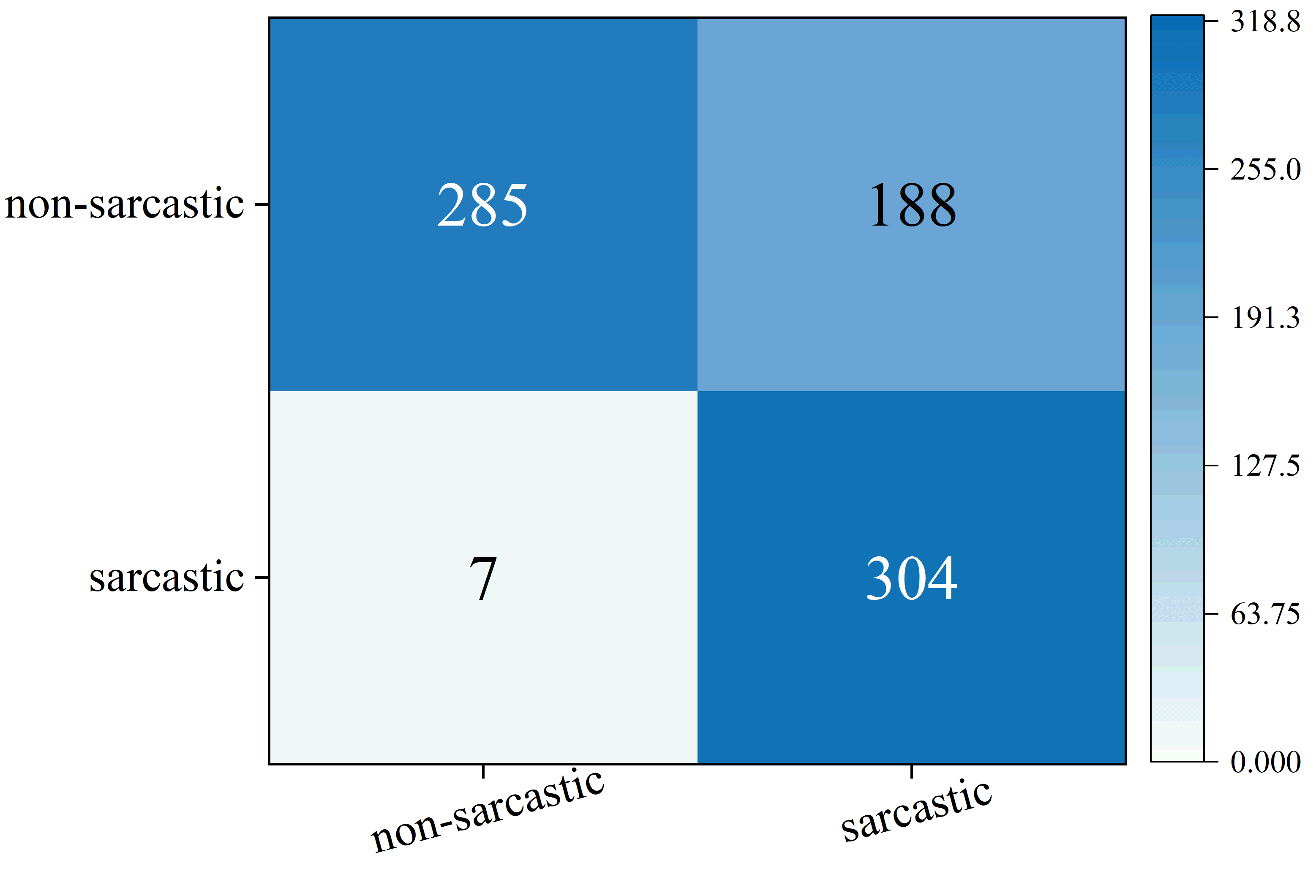}
    \label{fig: SinglePrompt}}
    \hfil
    \subfloat[Ghosh]{\includegraphics[width=2.1in]{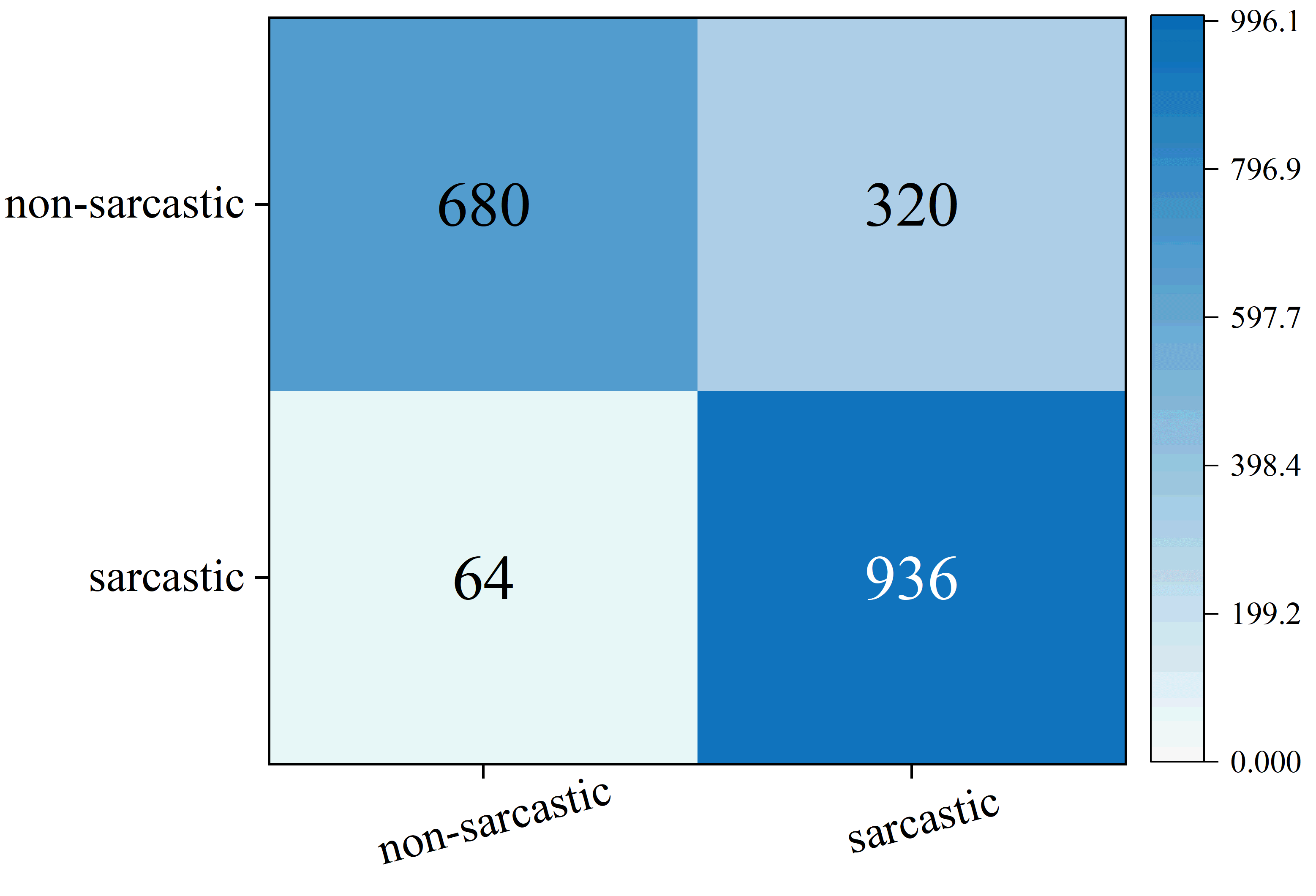}
    \label{fig: SinglePrompt}}
    \hfil 
\caption{The normalized confusion matrices for GPT-4 across six datasets in the \textbf{few-shot CoT setting}.}
\label{fig:confusionmatrix3}
\end{figure*}

\subsection{Error Analysis}\label{erroranalysis}
The detailed error analysis is also conducted via the confusion matrices that are shown in Fig.~\ref{fig:confusionmatrix1}, Fig.~\ref{fig:confusionmatrix2} and Fig.~\ref{fig:confusionmatrix3}. Each cell $\left ( i,j \right ) $ represents the percentage of class $i$ that is classified to be class $j$. Upon reviewing the classification results produced by GPT-4 Turbo on six datasets with three prompting approaches, we discover that imbalanced categories and the sarcastic samples are the key factors contributing to misclassification. 

In the zero-shot IO setting, GPT-4 Turbo performs well in identifying non-sarcastic instances, particularly on datasets like IAC-V2 and SemEval Task 3, where the true positive rates for non-sarcastic labels are very high. However, the model struggles more with distinguishing sarcastic instances, especially on datasets like IAC-V1 and iSarcasmEval, where the true positive rates for sarcastic labels are lower (0.46 and 0.38, respectively). This indicates that without any prior examples, the model tends to lean towards non-sarcastic predictions, failing to capture the nuanced language that often accompanies sarcasm. 

In the few-shot IO setting, we observe the improvement in performance across the datasets. The model benefits from a few examples, particularly in non-sarcastic classifications. In datasets like IAC-V2, the model still struggles, with only 46\% of sarcastic instances correctly identified. The model shows improvement but remains inconsistent in sarcastic classifications, indicating that while few-shot prompting provides some benefit, it may not be sufficient to fully capture sarcasm's complexity.

In the few-shot CoT setting, GPT-4 Turbo also similar difficulties in distinguishing between sarcastic and non-sarcastic instances, particularly in datasets with more subtle or nuanced sarcastic content.The reason is that CoT is not suitable to solve human sarcasm understanding tasks.

The improvements from few-shot IO and CoT are present but limited, suggesting that sarcasm requires more sophisticated handling, potentially incorporating richer contextual understanding and more advanced prompting strategies.

\subsection{Case Study}\label{sec:casestu}
Table~\ref{tab:casestudy} presents several examples where three representative LLMs made predictions. We conduct a brief analysis of the reasons behind these prediction errors. 

\textbf{(1) Contextual misunderstanding.} GPT-4 struggled with texts where sarcasm was deeply tied to contextual or cultural nuances. For instance, in Example 3 (``Being half Spanish and not being able to speak Spanish is honestly so disappointing''), the sarcasm lies in the speaker's ironic expression of disappointment. GPT-4 misclassified this as non-sarcastic, possibly due to its difficulty in detecting sarcasm embedded in personal or cultural identity issues. Similarly, in Example 4 (``I literally just ate 1/4 of a pan of brownies ... \#damnit \#notagain''), the self-deprecating tone and hashtags signal sarcasm, which GPT-4 failed to recognize.

\textbf{(2) Literal interpretation.} Claude 3 and LLaMA 3 generally performed better in identifying sarcasm when there was a stark contrast between the literal meaning and the intended message. For example, in Example 1 (``So many useless classes, great to be a student''), both models correctly identified the sarcastic tone. However, Claude 3 failed to detect sarcasm in Example 5 (``i love doing laundry''), where the lack of overt emotional indicators may have led to a more literal interpretation, causing the model to misclassify it as non-sarcastic.

\textbf{(3) Bias.} All three models misclassified Example 6 (``I just love not hanging out with my boyfriend'') as sarcastic, despite the straightforward nature of the statement. This suggests that the models might be overfitting to common sarcastic patterns, such as the phrase ``I just love...'' which is often used sarcastically. This shows that the models rely heavily on pattern recognition rather than contextual understanding. In addition, the ambiguity of the phrase posed a challenge for Claude 3 and LLaMA 3.

This analysis underscores the necessity for LLMs to develop a more profound contextual comprehension and enhanced interpretative skills to minimize prediction inaccuracies in sarcasm detection.

\begin{table*}[ht!]
\centering
\footnotesize
\caption{Typical examples for case study.} 
\label{tab:casestudy}
\scalebox{1}{
\begin{tabular}{clcccc}
\toprule
\textbf{Example} & \multicolumn{1}{c}{\textbf{Text}}                                                   & \textbf{Golden} & \textbf{GPT-4} & \textbf{Claude 3} & \textbf{LLaMA3} \\ \toprule
1                & \noindent \colorbox{gray!13}{\parbox{8.2cm}{So many useless classes, great to be student.}}                                       & sarcastic       & √                        & √                       & √                  \\
2                & \noindent \colorbox{gray!13}{\parbox{8.2cm}{I just love having grungy ass hair \#not.}}                                         & sarcastic       & √                        & √                       & √                  \\
3                & \noindent \colorbox{gray!13}{\parbox{8.2cm}{Being half spanish and not being able to speak spanish is honestly so disappointing.}} & sarcastic       & ×                        & √                       & √                  \\
4                & \noindent \colorbox{gray!13}{\parbox{8.2cm}{I literally just ate 1/4 of a pan of brownies ... \#damnit \#notagain.}}               & sarcastic       & ×                        & √                       & √                  \\
5                & \noindent \colorbox{gray!13}{\parbox{8.2cm}{i love doing laundry.                                                               }} & sarcastic       & √                        & ×                       & √                  \\
6                & \noindent \colorbox{gray!13}{\parbox{8.2cm}{I just love not hanging out with my boyfriend.}}                                     & non-sarcastic   & ×                        & ×                       & ×                  \\
7                & \noindent \colorbox{gray!13}{\parbox{8.2cm}{10 tell us how you really feel.}}                                                      & non-sarcastic   & √                        & ×                       & ×                  \\ \toprule
\end{tabular}
}
\end{table*}

\subsection{Evaluating LLMs on Multi-Modal Sarcasm Understanding}
Due to the rapid surge of multi-modal data on social networks, leveraging multi-modal information to enhance human language understanding and reasoning has become increasingly attractive and significant. Beyond text-based LLMs, multi-modal LLMs typically incorporate modality encoders, connectors, and generators. Through modality-aligned pre-training, these models acquire the ability to process multi-modal information. Therefore, in this section, we aim to investigate their capabilities for understanding multi-modal sarcasm.

More specifically, we select two popular and publicly accessible multi-modal sarcasm detection datasets, namely, MMSD~\cite{cai-etal-2019-multi} and CMMA~\cite{zhang2024cmma}, as our experimental platforms. We evaluate four state-of-the-art multi-modal LLMs: Retrieval-LLaVA 1.5~\cite{tang2024leveraging}, GPT-4V\footnote{https://openai.com/index/gpt-4v-system-card/}, Wenxin 4\footnote{https://yiyan.baidu.com/}, and Qwen-VL-Plus~\cite{bai2023qwenvlversatilevisionlanguagemodel}. Given that CoT prompting is less effective in  sarcasm understanding, we employ only the zero-shot IO prompting method for GPT-4V, Qwen-VL-Plus and Wenxin 4. 
Their prompts are: ``\textit{Based on the image and corresponding text below, analysis both the image and the text, determine whether they are sarcastic or not. If they are sarcastic output `sarcastic', otherwise, output `non-sarcastic'. Return the label only without any other text.}''
Retrieval-LLaVA 1.5 proposes a retrieval module for LLaVA 1.5 to search for demonstrations, aiming at further bridging the gap between LLaVA 1.5 and the specific multi-modal sarcasm detection task. 
Additionally, we present four state-of-the-art multi-modal sarcasm detection baselines for comparison. The results are shown in Table~\ref{tab:multimodalsarcasm}.

We can notice: (1) Zero-shot IO prompting-based multi-modal LLMs exhibit weak performance across both datasets when compared to PLMs. This suggests a possible limitation in their generalizability or adaptability to complex multi-modal tasks without additional contextual support. (2) Among the LLMs tested, GPT-4V shows superior results on the MMSD dataset, which is English, compared to Wenxin 4 and Qwen-VL-Plus.  In contrast, Qwen-VL-Plus excels on the CMMA dataset, which is Chinese. This divergence likely stems from the inherent differences in how sarcasm is expressed and contextualized in English versus Chinese. This indicates that GPT-4V may be  more suitable for processing sarcasm in English while Qwen-VL-Plus may be suited for Chinese sarcasm understanding. (3) The superior performance of Retrieval-LLaVA 1.5, particularly notable when it utilizes a retrieval approach to incorporate relevant examples into its prompts. This model outperforms the four PLMs significantly, demonstrating that adding contextually appropriate examples can substantially enhance model performance. We also present several examples in Fig.~\ref{fig:multimodalsarcasmcase}.

\begin{table*}[t!]
\centering
\caption{Performance on two multi-modal sarcasm datasets.  \textbf{\textcolor{blue}{Blue}} indicates the best results across LLMs.}
\label{tab:multimodalsarcasm}
\scalebox{0.88}{
\begin{tabular}{cl|cccc|cccc}
\midrule[1pt]
                                                   &                         & \multicolumn{4}{c}{\textbf{MMSD}}                                     & \multicolumn{4}{c}{\textbf{CMMA}}                                                                                               \\ \cline{3-10}
                                                   
\multirow{-2}{*}{\textbf{Paradigm}}  & \multirow{-2}{*}{\textbf{Model}} & \textbf{Acc}          & \textbf{P}          & \textbf{R}           & \textbf{F1}            & \textbf{Acc}          & \textbf{P}          & \textbf{R}           & \textbf{F1}                      \\\midrule[1pt]

                            & MMMA~\cite{zhang2024cmma}                 & 86.7          & 80.2         & 84.4           & 82.6          & 77.5        & 76.3          & 74.2           & 75.2                \\

                             & ResNet+BERT~\cite{pan-etal-2020-modeling}                 & 84.8         & 77.8         & 84.1           & 80.8      & 69.4    & 67.8         & 65.7          & 66.8                          \\

                              & MILNet~\cite{Qiao_Jing_Song_Chen_Zhu_Nie_2023}                 & 89.5          & 85.2         & 89.2           & 87.1          & -         & -          & - & -             \\

\multirow{-4}{*}{Multi-Modal PLMs}             & Multi-view CLIP~\cite{qin-etal-2023-mmsd2}          & 88.3          & 82.7         & 88.7           & 85.6          & -         & -          & -           & -               \\
 \hline
 
                          &              Retrieval-LLaVA 1.5~\cite{tang2024leveraging}           & \textbf{\textcolor{blue}{\underline{90.0}}}          & \textbf{\textcolor{blue}{\underline{89.3}}}         & \textbf{\textcolor{blue}{\underline{89.6}}}           & \textbf{\textcolor{blue}{\underline{89.4}}}          & -         & -          & -           & -              \\

      &    Wenxin 4            & 60.0          & 52.6         & 31.5           & 39.4          & 50.0 & \textbf{\textcolor{blue}{\underline{40.0}}}          & 50.0           & 44.4                 \\

      &    Qwen-VL-Plus             & 64.4         & 52.7         & 95.0          & 67.8          & 39.0 & 39.4          & \textbf{\textcolor{blue}{\underline{97.5}}}           & \textbf{\textcolor{blue}{\underline{56.1}}}                \\

\multirow{-4}{*}{Multi-Modal LLMs}    &  GPT-4V           & 71.0          & 59.4         & 96.5 & 73.5          & \textbf{\textcolor{blue}{\underline{52.0}}}         & 31.8          & 17.6           & 22.6               \\
\midrule[1pt]

\end{tabular}
}
\end{table*}

\begin{table*}[t!]
\centering
\caption{Typical multi-modal examples for case study.}
\label{fig:multimodalsarcasmcase}
\footnotesize
\scalebox{0.8}{
\begin{tabular}{cclcccc}
\hline
\textbf{Example} & \textbf{Image} & \textbf{Text}                                          & \textbf{Golden} & \textbf{GPT-4V} & \textbf{Qwen-VL} & \textbf{WenXin4} \\ \hline
1                & \includegraphics[width=2cm]{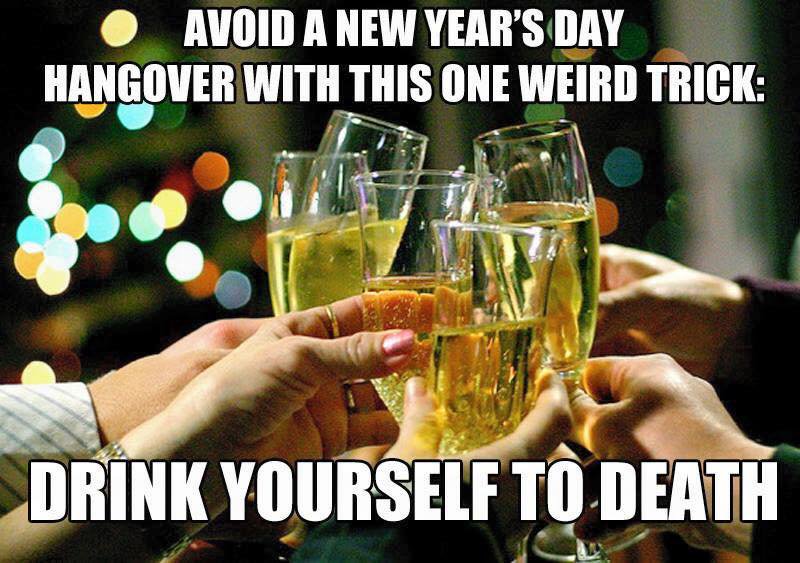} & \noindent \colorbox{gray!13}{\parbox{5.2cm}{happy new year , everyone ! xoxo \# year2016 \# partyhard \# nihilistmemes}} & sarcastic       & √               & √                & ×                \\
2                & \includegraphics[width=2cm]{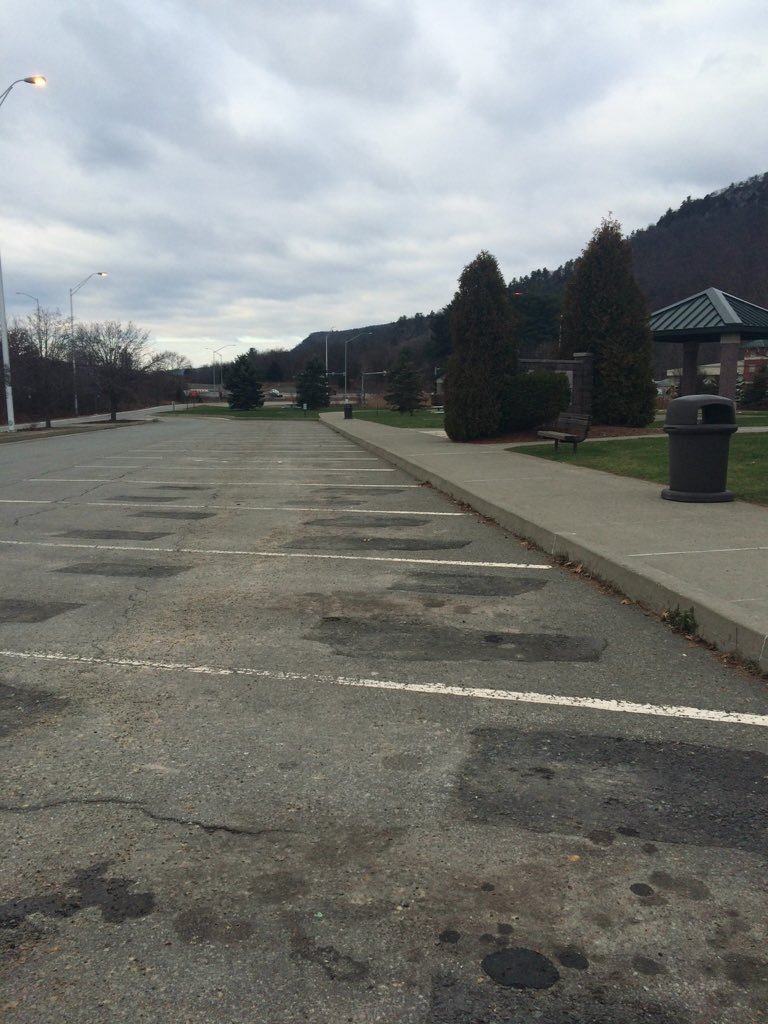} & \noindent \colorbox{gray!13}{\parbox{5.2cm}{the pa welcome center is hopping today .}}                                   & sarcastic       & √               & √                & ×                \\
3                & \includegraphics[width=2cm]{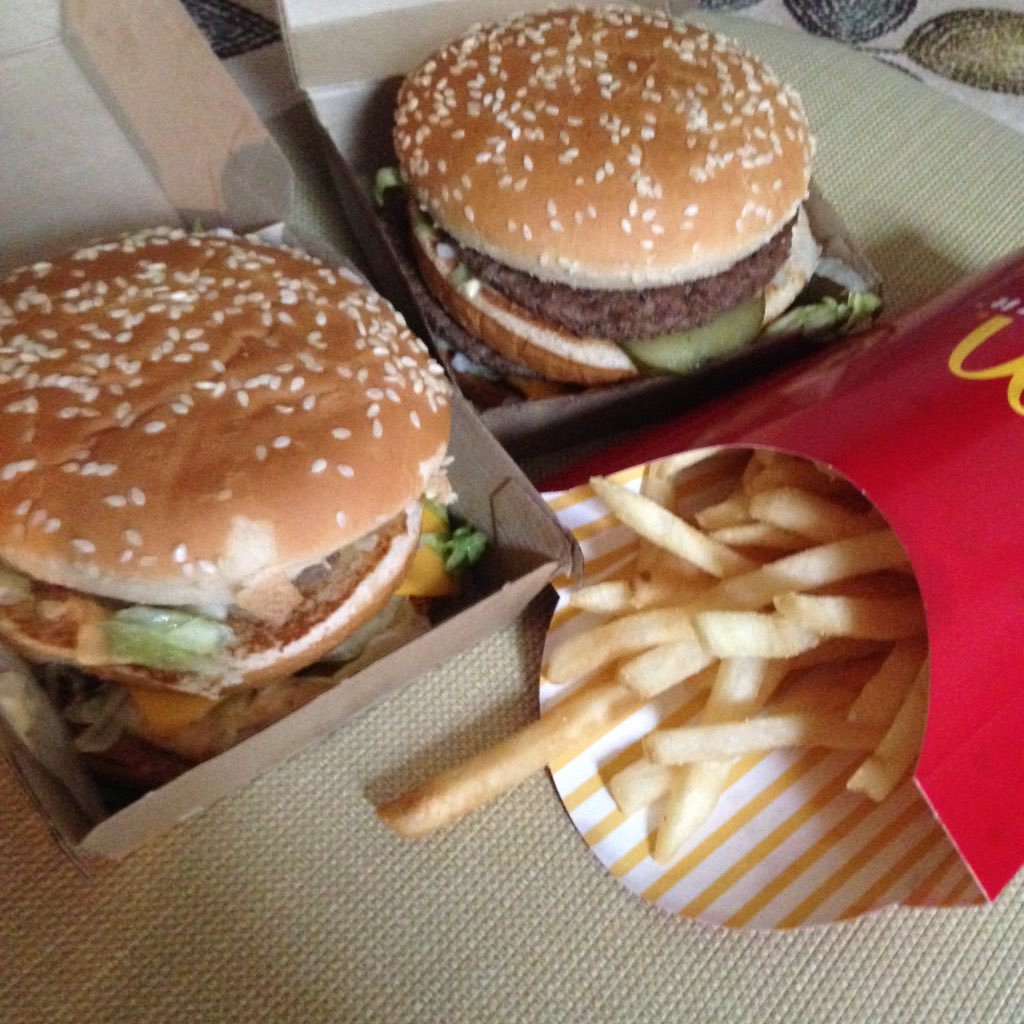} &  \noindent \colorbox{gray!13}{\parbox{5.2cm}{feeding my abs nothing but the best quality beef}}                           & sarcastic       & √               & √                & ×                \\

4                & \includegraphics[width=2cm]{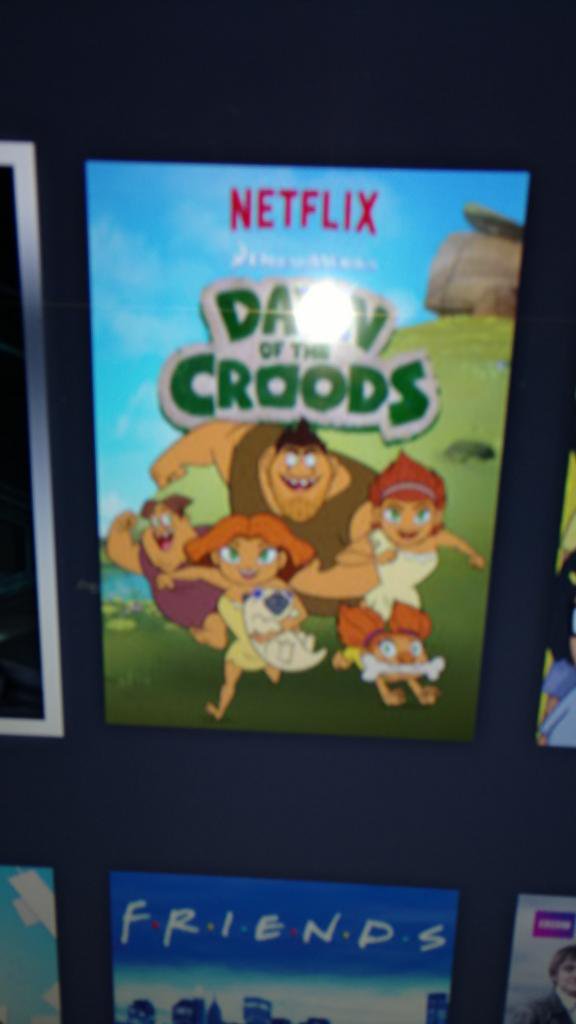} &  \noindent \colorbox{gray!13}{\parbox{5.2cm}{because every mildly successful cgi film needs an animated spinoff .}}       & sarcastic       & √               & √                & ×                \\
5                & \includegraphics[width=2cm, keepaspectratio]{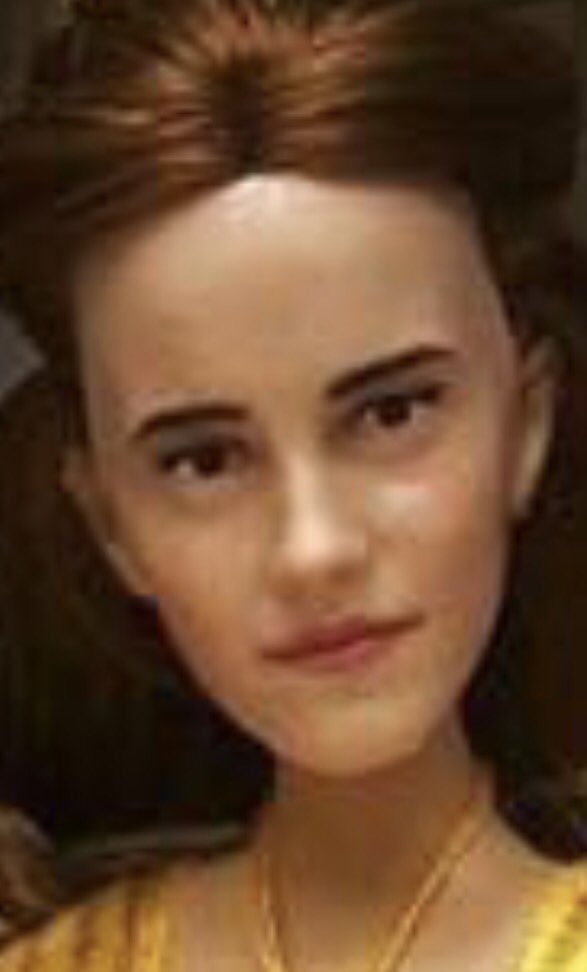} &   \noindent \colorbox{gray!13}{\parbox{5.2cm}{when ryan seacrest air-kisses you , but you went for the real kiss}}         & non-sarcastic   & ×               & ×                & ×                \\
6                & \includegraphics[width=2cm, keepaspectratio]{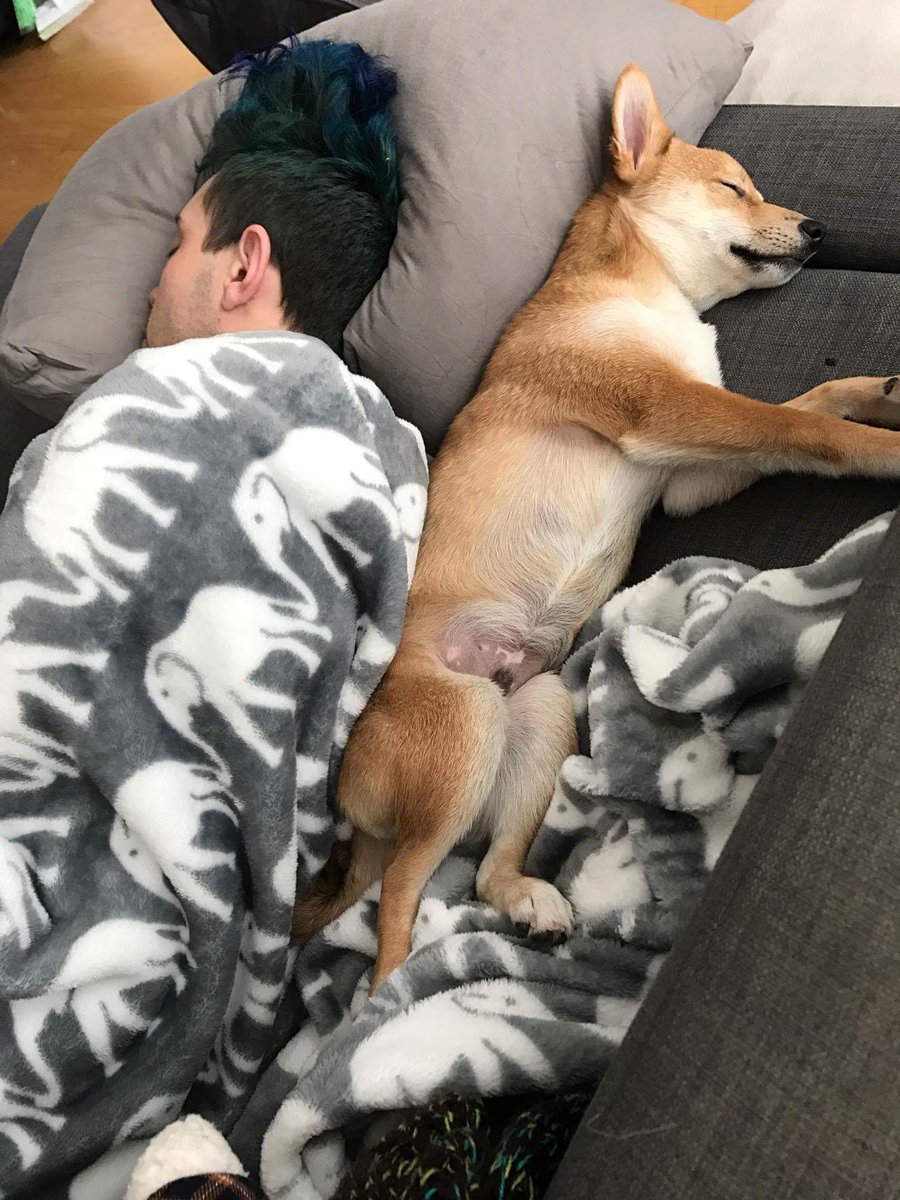} & \noindent \colorbox{gray!13}{\parbox{5.2cm}{bae and i ...... all day .}}                                                 & non-sarcastic   & √               & √                & √                \\ \hline
\end{tabular}
}
\end{table*}

\section{Conclusion}\label{con}
In the era of large language models (LLMs), there is growing concern that LLMs' success may not be fully tenable when considering sarcasm understanding. 
To address this question, we select eleven SOTA LLMs and eight SOTA PLMs and present comprehensive evaluations on six widely used benchmark datasets.
The results show that current LLMs underperform supervised PLMs based sarcasm detection baselines across six sarcasm benchmarks. In addition, GPT-4 consistently and significantly outperforms other LLMs across various prompting methods. This suggests that significant efforts are still required to improve LLMs' understanding of human sarcasm. In the future, we plan to improve the classification performance by integrating the predictions of multiple models using methods such as bagging or boosting.

\textbf{Limitations.} 
Our research also has several limitations. (1) We only employed standard prompting methods and chain-of-thought techniques to guide large language models in detecting human sarcasm, without exploring more refined prompting methods such as tree-based or graphical approaches. These methods will be further investigated in our future work. (2) Our study did not consider the variation in model performance on sarcasm across different contexts. Future work could explore how to enhance the model's understanding of sarcastic language by refining context processing mechanisms. 
\end{CJK}
\bibliography{nn}
\end{document}